\documentclass[10pt]{article} % For LaTeX2e
\usepackage[preprint]{tmlr}
\usepackage{hyperref}
\usepackage{url}
\usepackage{amssymb}
\usepackage{amsmath}
\usepackage{algorithm}
\usepackage[noend]{algpseudocode}
 \usepackage{multirow}
 \usepackage{subfig}
\usepackage{graphicx}
\usepackage{lscape}
\usepackage{xcolor}
\usepackage{booktabs}
\usepackage{threeparttable}
\usepackage{graphicx}   % for \resizebox
\usepackage{multirow}

\title{Uncertainty Quantification in SVM prediction}

\author{\name Pritam Anand \email pritam$\_$anand@daiict.ac.in \\
      \addr DA-IICT, Gandhinagar.
     }

\begin{document}

\maketitle

\begin{abstract}
This paper explores Uncertainty Quantification (UQ) in SVM predictions, particularly for regression and forecasting tasks. Unlike the Neural Network, the SVM solutions are typically more stable, sparse, optimal and interpretable. However, there are only  few literature which addresses the UQ in SVM prediction. At first, we provide a comprehensive summary of existing Prediction Interval (PI) estimation and probabilistic forecasting methods developed in the SVM framework and evaluate them against the key properties expected from an ideal PI model. We find that none of the existing SVM PI models achieves a sparse solution, which has remained a key advantage of the standard SVM model developed for classification and regression tasks.  To introduce sparsity in SVM model, we propose the Sparse Support Vector Quantile Regression (SSVQR) model, which constructs PIs and probabilistic forecasts by solving a pair of linear programs. Further, we develop a feature selection algorithm for PI estimation using SSVQR that effectively eliminates a significant number of features while improving PI quality in case of high-dimensional dataset. Finally we extend the SVM models in Conformal Regression setting for obtaining more stable prediction set with finite test set guarantees.  Extensive experiments on artificial, real-world benchmark datasets compare the different characteristics of both existing and proposed SVM-based PI estimation methods and also highlight the advantages of the feature selection in PI estimation.  Furthermore, we compare both, the existing and proposed SVM-based  PI estimation models, with modern deep learning models for probabilistic forecasting tasks on benchmark datasets. Furthermore, SVM models show comparable or superior performance to modern complex deep learning models for probabilistic forecasting task in our experiments.The code are available at $\mbox{https://github.com/ltpritamanand/-PI-IN-SVM-}$.

\end{abstract}

\section{Introduction}
\label{sec:sample1}
Given the training set $T=\{(x_i,y_i): x_i \in \mathbf{R}^n, y_i \in \mathbf{R}, i =1,2,...m. \}$, sampled independently from the joint distribution of the random variables $(X,Y)$, the goal of the regression task is to estimate a function that predicts the target variable  $y$ based on the input variable $x$ well. However, in most of applications, the prediction of the regression model may not be perfectly accurate due to random relationship between $Y$ and $X$. For example, predicting the impact of a specific drug on a patient's heart rate based on their Body Mass Index (BMI) may  not be accurate and  may involve a significant degree of uncertainty. In such cases, quantifying these uncertainties is crucial for making effective decisions.

 The Prediction Interval (PI) estimation is most commonly used Uncertainty Quantification (UQ) technique in regression tasks. Given a high confidence $1-\alpha \in (0,1)$ and training set $T$, the PI  tube is defined as a pair of functions $(f_1,f_2)$. It is said to be well calibrated if it satisfies $P(f_1(X)\leq Y \leq f_2(X)|X) \geq 1-\alpha$. The objective of the PI models is to obtain a PI tube with the minimum possible width while ensuring the target calibration. Therefore, the performance of a PI estimation method is mainly evaluated using two criteria: Prediction Interval Coverage Probability (PICP), which computes the fraction of $y$ values within the PI tube, and Mean Prediction Interval Width (MPIW), quantifying the width of the PI tube.

The  PI estimation models requires to explore the different  characteristics of the conditional distribution $Y|X$, rather focusing only on $E(Y|X)$ as done in standard regression tasks.  The basic approach of PI models involves estimating a pair of quantile functions (\cite{koenker1978regression}), say ($f_{q}(x),f_{1+q-\alpha}(x)$), of the  conditional distribution $Y|X$, for some $0 \leq q \leq \alpha$, where the $q^{th}$ quantile function for given $x$ is defined as infimum of functions satisfying $P(y\leq f_q(x)|x) = q$.  

For time-series data, estimating the Prediction Interval (PI) for future observations using an auto-regressive approach is referred to as probabilistic forecasting.
 Both PI estimation and probabilistic forecasting models are widely investigated in the Neural Network (NN) architectures in the literature. The PI estimation methods in the NN literature can primarily be divided into two main categories. A popular class of PI estimation methods assumes that the conditional distribution $Y|X$ follows a particular distribution (often normal) and obtains the quantile functions by computing the inverse of the  corresponding cumulative distribution function. Some important of them are Bayesian method (\cite{mackay1992evidence, bishop1995neural}, Delta method\cite{de1998prediction,hwang1997prediction,seber2015nonlinear}) and Mean Variation Estimation (MVE) method (\cite{nix1994estimating}). Some of the recent NN architecture for the probabilistic forecasting task with distribution assumptions are Mix Density Network (\cite{bishop1994mixture,zhang2020improved}), Deep Auto-regressive Network (Deep AR) (\cite{salinas2020deepar}).

The other class of PI estimation and probabilistic forecasting methods believe in estimating the  pair of quantile functions ($f_{q}(x),f_{1+q-\alpha}(x)$) in distribution-free setting without imposing any assumption regarding the distribution of $Y|X$. For estimation of the  $q^{th}$ quantile function, $f_q(x)$, most of them minimizes the  pinball loss function.  The pinball loss-based NN model, also known as Quantile Regression Neural Network (QRNN) (\cite{taylor2000quantile,cannon2011quantile}) is the main PI estimation method, which has been utilized in various engineering applications. The pinball loss-based NN model has been frequently applied to probabilistic forecasting of wind (\cite{wan2016direct}),  electric load (\cite{zhang2018improved,zhang2020improving}), electric consumption (\cite{he2019electricity}), flood risks (\cite{pasche2024neural}) and solar energy (\cite{lauret2017probabilistic}). 
Some of the distribution-free PI estimation  NN methods consider the minimization of a particularly designed loss function for the direct and simultaneous estimation of the bounds of the PI. Some of them are Lower Upper Bound Estimation (LUBE) NN, Quality-Driven (QD) Loss NN (\cite{pearce2018high}) and Tube loss NN (\cite{anand2024tube}).

However, a well-calibrated HQ PI guarantees the target coverage level $t$ only asymptotically, and may fail to achieve it on finite test samples. In real-world decision-making, especially in high-stakes applications, guarantees on finite test samples coverage are often essential. Conformal Regression (CR) (\cite{vovk1999machine,vovk2005algorithmic}) provides a principled framework through which PI models can be adapted to ensure such finite-sample coverage guarantees, making them more suitable for practical deployment.

Despite the remarkable success of neural architectures, researchers still prefer SVMs for their predictive accuracy in regression tasks, particularly when dealing with small size tabular dataset. This is because SVM regression models explicitly incorporate regularization and most of them minimize a convex program to guarantee a global optimal, interpretable and sparse solutions, which remain missing in the NN learning. 

Compared to SVM models, NN and deep learning-based regression models typically exhibit a higher degree of model uncertainty. This arises for two main reasons. First, neural models generally have a much larger number of parameters than SVMs, making them more prone to  variability. Second, due to the non-convex nature of their optimization landscape, NN models often converge to different local minima across different training trails, even under identical training set and hyperparameter settings. 

 However, in contrast to NN literature, there are only a few SVM methods which target the PI estimation and probabilistic forecasting tasks in the literature. Moreover, SVM models have been largely unexplored within conformal regression setting. Our work addresses these gaps in the literature by extending contemporary UQ techniques to the SVM framework, supported by a comprehensive analysis and comparisons.  We summarize the contribution of our work in detail as follows. 
\begin{enumerate}
    \item [(a)] First, we carefully review the existing literature on PI estimation and probabilistic forecasting methods in SVM. We outline the desirable properties of an ideal PI model and compare the PI estimation and probabilistic forecasting methods in SVM against them.  We find that only two of SVM PI methods attain the global optimal solution but, none of them achieve a sparse solution vector. 

    \item[(b)] Building on this motivation, we propose a sparse SVM method for Prediction Interval (PI) estimation and probabilistic forecasting.  Our Sparse SVM model enhances the PI estimation process by reducing the overall complexity of learning and prediction while preserving the classical properties of SVM, achieving both a globally optimal and sparse solution. Further, we highlight the importance of the feature selection in PI estimation particularly in high dimensional regression tasks. For this, we develop a simple yet effective feature selection algorithm for PI estimation using our sparse SVM PI model. We show that our algorithm does not only successfully discard a significant percentage of features but, also improves the quality of the PI  while learning the PI for high-dimensional data. To the best of our knowledge, there is no any existing literature which study the feature selection problem in context of PI estimation.

    %For a given quantile $q \in (0,1)$ and target calibration level $1-\alpha$, the method computes two sparse quantile estimates, $(\hat{F}_{q}(x), \hat{F}_{1+q-\alpha}(x))$, which serve as the bounds of the PI. To achieve sparsity in the solution for $\hat{F}_{q}(x)$ and $\hat{F}_{1+q-\alpha}(x)$, the method optimizes the regularization of the norm $L_1$ alongside the pinball loss function parameterized by $q$ and $(1+q-\alpha)$ in two separate problems, each of them efficiently formulated as a linear programming problem. Consequently, the proposed approach solves a pair of linear programming problems to derive the sparse SVM solution for PI estimation and probabilistic forecasting tasks.

    \item [(c)]  Finally, we extend SVM regression to the conformal regression setting to achieve finite-sample coverage guarantees. Compared to NN, we show that the conformal prediction sets produced by SVM models  
 are more stable and interpretable due to its global optimal solution.

    %We develop a simple yet effective feature selection algorithm for PI estimation using our sparse SVM PI model. We show that our algorithm does not only successfully discard a significant percentage of features but, also improves the quality of the PI  while learning the PI for high-dimensional data. To the best of our knowledge, our algorithm is the first to address the  feature selection problem in PI estimation for tabular data, paving the way for more efficient and interpretable UQ models for quantifying the uncertainty of prediction with high dimensional data.   
    
  \item [(d)]  We conduct extensive experiments on artificial, real-world benchmark datasets to empirically analyze the PI quality obtained by the both existing and proposed PI estimation models in SVM. For high-dimensional datasets, we reveal the effectiveness of the Sparse SVM-based PI model by performing feature selection using sparse SVM PI based feature selection algorithm. Our numerical result also demonstrate that the SVM based probabilistic forecasting models can achieve comparable, and in some cases superior, PI quality relative to recent complex deep probabilistic forecasting  models that involve significantly larger numbers of parameters on few  benchmark datasets.

    %superiority of the proposed Sparse SVM model over existing SVM models for the PI estimation task. Additionally, we applied the Sparse SVM model to obtain the probabilistic forecast of wind speed  and showed that it achieves superior PI quality compared to other SVM models used for probabilistic forecasting.
\end{enumerate}

The remainder of this paper is structured as follows. Section 2 provides a systematic review of the prerequisite concepts required for understanding of the SVM models and UQ techniques.  Section-3 provides a detail description of the several SVM models for PI estimation and probabilistic forecasting,   highlighting their advantages and limitations. In Section 4, we introduce the proposed Sparse SVM models for PI estimation, probabilistic forecasting and conformal regression tasks. Section 5 presents the numerical results from extensive experiments, demonstrating the effectiveness of the proposed SVM model for PI estimation and probabilistic forecasting tasks. Section 6 outlines the future work.

\section{Related key concepts}
In this section, we will outline key concepts and methods relevant to PI estimation techniques in SVM.  
\subsection{Quantile Regression and SVM}
\begin{figure}[h]
    \centering
    \includegraphics[width=0.4\linewidth]{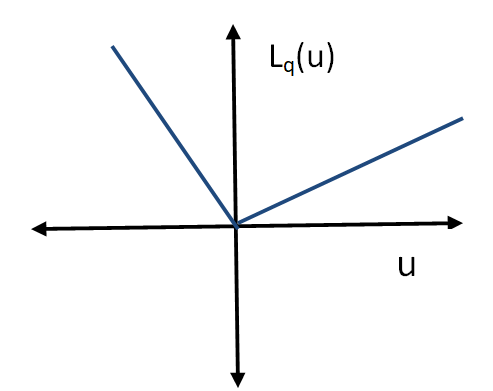}
    \caption{Pinball loss function}
    \label{fig:enter-label}
\end{figure}
In distribution free setting, for a given quantile $q \in (0,1)$, the quantile value is estimated by minimizing the pinball loss function, which is given by 
\begin{equation}
     \rho_q(u)  = \begin{cases}
        qu, \mbox{ ~~~if~~} u \geq 0, \\
        (q-1)u, \mbox{Otherwise.}, 
    \end{cases} 
\end{equation}
For the estimation of the conditional quantile function,  
 $u$ represents the error obtained by the subtracting the estimates $f(x_i)$ from its target values $y_i$.  For a given quantile $q \in (0,1)$, training set $T=\{(x_i,y_i): x_i \in \mathbf{R}^n, y_i \in \mathbf{R}, i =1,2,...m. \}$ and class of function $F$, let us suppose that $f_T$ is the solution of the problem $\min \limits_{f\ \in F} \sum \limits_{i=1}^{m}\rho_q(y-f(x_i))$. Tekeuchi et al., have shown that the fraction of $y$ values lying below the function $f_T(x)$ is bounded from above by $qm$ and asymptotically equals $qm$ with probability $1$ under a very general condition in their work (\cite{takeuchi2006nonparametric}).

 Given the training set, SVM models estimate the function in the form of $f(x) = w^T\phi(x) + b$, where $\phi$ maps the input variable $x$ into the high dimensional feature space, such that for any pair of $x_i$ and $x_j$ in   $\mathbf{R}^n$,  $\phi(x_i)^T\phi(x_j)$ can be obtained by the well defined kernel function $k(x_i,x_j)$. By the use of the kernel trick and representer theorem \cite{scholkopf2001generalized},  the SVM estimate $f(x) = w^T\phi(x) + b$ can be represented by the kernel generated function in the form of $\sum_{i=1}^{m}k(x_i,x)u_i + b$, where $k$ is positive-semi definite kernel \cite{mercer1909xvi}. This representation eliminates the need for explicit knowledge of the mapping $\phi$.

 \subsection{Support Vector Quantile Regression model}
The Support Vector Quantile Regression (SVQR) model minimizes the $L_2$-norm of the regularization along with the  empirical risk computed by the pinball loss function.  For $q^{th}$ quantile function estimation, it seeks the solution of the problem
\begin{equation}
    \min \limits_{(w,b)} \frac{1}{2} w^Tw + C\sum \limits_{i=1}^{m} \rho_q(y_i- (w^T\phi(x_i) +b)),
\end{equation}
which can be equivalently converted to the following Quadratic Programming Problem (QPP)
 \begin{eqnarray}
     \min \limits_{(w,b,\xi,\xi^*)} \frac{1}{2} w^Tw + C\sum \limits_{i=1}^{m} (q\xi_i + (1-q)\xi_i^{*}) \nonumber \\
     & \hspace{-110mm}\mbox{subject to,} \nonumber \\
     & \hspace{-80mm} y_i- (w^T\phi(x_i) +b) \leq \xi_i, \nonumber  \\
     & \hspace{-80mm} (w^T\phi(x_i) +b)-y_i \leq \xi_i^{*}, \nonumber \\
     & \hspace{-80mm} \xi_i, \xi_i^{*} \geq 0,~ i =~1,2,..m,
     \label{eq1}
 \end{eqnarray}
where $C \geq 0$ is the user defined parameter for trading-off the empirical risk against the model complexity. 

To efficiently solve QPP (\ref{eq1}), we often focus on obtaining the solution to its corresponding Wolfe dual problem, which is given by 
\begin{eqnarray}
     \min \limits_{(\alpha ,\beta)} \sum_{i=1}^{m} \sum_{j=1}^{m}(\alpha_i-\beta_i)k(x_i,x_j)(\alpha_j-\beta_j) - \sum_{i=1}^{m}(\alpha_i-\beta_i)y_i \nonumber \\
     & \hspace{-180mm}\mbox{subject to,} \nonumber \\
     & \hspace{-140mm}  \sum \limits_{i=1}^{m}(\alpha_i-\beta_i) = 0, \nonumber  \\
     & \hspace{-120mm}  0 \leq \alpha _i \leq Cq, ~ i =~1,2,..m, \nonumber \\
     & \hspace{-110mm} 0 \leq \beta _i \leq C(1-q),~ i =~1,2,..m,
     \label{eq2}
 \end{eqnarray}
where $(\alpha_i, \beta_i), i= ~1,2,..,m$, are Lagrangian multipliers.

After obtaining the optimal solution of the dual problem (\ref{eq2}), $(\alpha_i^{*}, \beta_i^{*}), i= ~1,2,..,m$, the $q^{th}$ quantile function is estimated by 
\begin{equation}
    f_q(x) = \sum_{i=1}^{m}(\alpha_i^{*}- \beta_i^{*})k(x_i,x) + b . \label{out1}
\end{equation}
The estimation of the bias term $b$ can be obtained by using the  KKT conditions of the primal problem (\ref{eq1}). For this, we need to select the every training point $(x_k,y_k)$ which corresponds to $0 < \alpha_k^{*} < Cq$ or $0 < \beta_k^{*} < C(1-q)$ and compute
\begin{equation}
    b_k= y_k - \sum \limits_{i=1}^{m}(\alpha_i^{*}-\beta_i^{*}) k(x_i,x_k).
\end{equation}
The final value of bias $b$ can be obtained by computing the the mean of  all $b_k$. 

\subsection{Least Squares Support Vector Regression }
For estimating the mean regression using training set $T$, the LS-SVR model \cite{suykens2002weighted} minimizes the least square loss function along with the $L_2$-norm of regularization in the following problem.
  \begin{eqnarray}
     \min \limits_{(w,b,\xi)} \frac{1}{2} w^Tw + C\sum \limits_{i=1}^{m} (\xi^2_i) \nonumber \\
     & \hspace{-90mm}\mbox{subject to,} \nonumber \\
     & \hspace{-40mm} y_i- (w^T\phi(x) +b) = \xi_i,~ i =~1,2,..m.
     \label{eq4}
 \end{eqnarray}   
The solution of problem (\ref{eq4}) can be obtained through the  following system of equations
  \begin{equation}
  \begin{bmatrix}
  0 & e^T\\
  e & K(A,A^T) + \frac{2}{C}I
  \end{bmatrix} \begin{bmatrix}
  b \\
  \alpha
  \end{bmatrix} =
  \begin{bmatrix}
  0 \\
  Y
  \end{bmatrix}, \label{eq3}
  \end{equation}
where \( K(A, A^T) \) is an \( m \times m \) kernel matrix constructed from the training set \( T \), \( e \) is an \( m \)-dimensional column vector of ones, and \( I \) represents the \( m \times m \) identity matrix.
After obtaining the $(b,\alpha)$ from (\ref{eq3}), the LS-SVR estimates the regression function for a given $x \in \mathbb{R}^n$ using
  \begin{equation}
  {f}(x) = \sum_{i=1}^{m}k(x_i,x)\alpha_i + b . \label{eq5}
  \end{equation}
  
\subsection{Probabilistic Forecasting} 
The task of probabilistic forecasting is basically an extension of the PI estimation in an auto-regressive setting. Consider the time series observations $  T = \{ x_1,x_2,....,x_t \}$, recorded on $t$ different time stamps.  If \( p < t \) denotes the effective lag window, then auto-regressive models estimate the relationship between \( z_i := (x_{i-p+1}, \ldots, x_i) \) and \( x_{i+1} \) for \( i = p, p+1, \ldots, t-1 \) using the training set \( T \), and use this learned relationship to forecast future observations. Point forecasting models aim to estimate the conditional expectation $E(x_{i+1} \mid z_i)$. However, in many cases, such forecasts may incur significant errors due to inherent noise and volatility in the data. Probabilistic forecasting quantifies these uncertainties in the prediction by obtaining the PI. 

 The task of probabilistic forecasting is to estimate the PI for $x_{i+1}$ given the input $z_{i}$ for $i \geq  t$. The SVM based probabilistic forecasting models obtain the estimate of  the PI $[\hat{F}_{q}(z_{i}),\hat{F}_{1+q-\alpha}(z_{i})]$, where $\hat{F}_{q}(z_{i})$ and $\hat{F}_{1+q-\alpha}(z_{i})$ are  kernel generated functions, estimating of the  $q^{th}$ and ${(1-\alpha-q)}^{th}$ quantiles of the conditional distribution $(x_{i+1} \mid z_{i})$ for some $ q \in (0,\alpha)$. Distribution-free probabilistic forecasting methods estimate quantile functions directly, without making any assumptions about the conditional distribution $(x_{i+1} \mid z_{i})$.

 \subsection{Conformal Regression}
 Conformal Regression (\cite{vovk1999machine,vovk2005algorithmic}) provides a general framework for adjusting PI models to guarantee the target coverage $1-\alpha$ on finite test samples, assuming only that the data are exchangeable. 

The split conformal regression (CR) approach (\cite{papadopoulos2002inductive,papadopoulos2008inductive})starts by dividing the available training data \( T \) into two separate subsets: a training set \( I_1 \) used to fit the predictive model, and a calibration set \( I_2 \) used for PI  adjustment. A nonconformity score function is then introduced to quantify the  disagreement between predicted value for \( y_i \), given input \( x_i \) and its actual observed value. These nonconformity scores are computed on the calibration set \( I_2 \). A obvious choice of the non-conformity score is the absolute residual value, computed by $|y_i-\hat{f}({x_i}) |, ~~  i \in I_2,$ where $\hat{f}$ is the estimate of the mean regression model trained on $I_1$. 

Romano et al. have developed  the quantile regression  based nonconformity score for obtaining the  fully adaptive conformal prediction set  in their work Conformalized Quantile Regression (CQR) (\cite{romano2019conformalized}),  which is given by
 \begin{equation}
     E_i = max \{\hat{F}_{lo}(x)-y_i, ~ y_i - \hat{F}_{hi}(x) \} , \label{nfs}
 \end{equation}
 where  $\hat{F}_{lo}(x)$ and   $\hat{F}_{hi}(x)$  are the estimates of the $q^{th}$ and  $(1+q-\alpha)^{th}$ quantile function on set $I_1$  for some $0 \leq q \leq \alpha$. 
 
After computing the non-conformity scores, the CR methodology requires the computation of $ (1-\alpha)(1+ \frac{1}{|I_2|})${-th empirical quantile of } the non-conformity score.  In case of CQR, we denote it with $Q_{1-\alpha}(E_i,I_2)$ and obtain the prediction set on new test point  $x_{m+1}$ as 
\begin{equation}
    C(x_{m+1}) = [ \hat{F}_{lo}({x_{m+1}}) - Q_{(1-\alpha)}(E,I_2) , \hat{F}_{hi}({x_{m+1}}) + Q_{1-\alpha}(E,I_2) ]  \label{crp}
\end{equation}

\section{PI estimation in SVM}
In this section, we gather in detail the PI estimation and probabilistic forecasting methods developed in the SVM literature and compare their advantages and limitations. 

\subsection{PI estimation through LS-SVR}
One of the naive PI estimation method in SVM literature follows the normal assumption regarding the distribution of $Y|X$ and estimate its mean through (\ref{eq5}) by training the LS-SVR model. The error distribution \(\epsilon_i = y_i - f(x_i)\) follows a normal distribution with a mean of zero and variance \(\sigma\). This variance can be estimated from the error computed on training set $T$. The pair of quantile bounds required for PI is estimated as $(\hat{f(x)}+\epsilon_{\frac{\alpha}{2}}, \hat{f(x)}+\epsilon_{1-\frac{\alpha}{2}}) $, where $\epsilon_q$ is the $q^{th}$ quantile of the error.
A more refined and bias-corrected PI based on the LS-SVR model is proposed in (\cite{de2010approximate,cheng2014confidence}).

\subsection{PI estimation through SVQR}
Given the high confidence $1-\alpha$ with training set $T$, the PI model requires the estimation of the pair of quantile functions ($f_{q}(x),f_{1+q-\alpha}(x)$) of the  conditional distribution $Y/X$. The SVQR model can be trained twice for the estimation of the  pair of these quantile functions for some $0 \leq q \leq \alpha$.
We detail the algorithm for PI estimation through SVQR in Algorithm 1. At Algorithm 1, the tuning of $C$ refers to selecting the value of $C$ from a specified range such that the SVQR estimate obtains the least coverage error.

\begin{algorithm}
\caption{PI estimation through SVQR}\label{alg:euclid}
\begin{algorithmic}[1]
\Procedure{:- PI through SVQR} {$T,1-\alpha$}
\State Choose some $\bar{q} \in (0,1-\alpha)$ 
\For{ each $q \in \{\bar{q}, (1+\bar{q}-\alpha)\} $} 
\State Solve the QPP problem (\ref{eq2}) by tuning the value of $C$. Obtain the solution $(\alpha^*,\beta^*)$.
\State  Estimate the function $f_q(x)$ using the (\ref{out1}). 
\EndFor\label{euclidendwhile}
\State \textbf{return} $(f_{\bar{q}}(x),f_{1+\bar{q}-\alpha}(x))$
\EndProcedure
\end{algorithmic}
\end{algorithm}

A key challenge in estimating prediction intervals (PI) using the quantile approach is to determine the good choice of $\bar{q}$  for obtaining the narrower PI. For a symmetric noise distribution, \(\bar{q} = \frac{\alpha}{2}\) is expected to produce the PI with minimum width. However, this does not hold for an asymmetric noise distribution. In the latter case, \(\bar{q}\) should be selected such that the resulting PI passes through the denser regions of the data cloud. Furthermore, for each choice of $\bar{q}$, the optimization problem (\ref{eq2}) must be solved twice for $q = \bar{q}$ and  $q = 1 + \bar{q} - \alpha$ to obtain the prediction interval (PI). It increases the overall computational complexity of the PI estimation process, making it both time-consuming and challenging in practice.

To simplify the PI estimation process, researchers have developed direct PI estimation methods, which solve a single optimization problem to  obtain the both bounds of PI simultaneously. These methods are designed with a specialized loss function that can be minimized to obtain the both bounds of PI simultaneously.  Some important of them are LUBE loss (\cite{khosravi2010lower}), Quality-Driven (QD) loss (\cite{pearce2018high}) and Tube loss (\cite{anand2024tube}) functions. We describe those also formulated within the SVM framework as follows.

\subsection{PI estimation through Tube loss}
(\cite{anand2024tube}) have developed the Tube loss for PI estimation and probabilistic forecasting. It can be minimized directly to obtain the bounds of the PI simultaneously. The minimizer of the Tube loss function also guarantees the target coverage $1-\alpha$ asymptotically.  Also, the PI tube can also be shifted up or down by tuning its parameter $r$ so that it can cross through the denser region of data cloud for minimal PI width. Furthermore, the width of the PI tube can be explicitly minimized in its optimization problem through the parameter $\delta$. It helps to improve the PI width,  when the PI tube achieves a coverage higher than the target on the validation set.

 The Tube loss function is a kind of two-dimensional extension of the pinball loss function (\cite{koenker1978regression}). For a given $1-\alpha \in (0,1)$ and $u_2 \leq u_1$,  the Tube loss function is given by
 \begin{eqnarray}
		  \rho_{1-\alpha}^{r}(u_2, u_1) =
		 \begin{cases}
                 (1-\alpha) u_2,  ~~\mbox{if}~~~ u_2  > 0,\\
			-\alpha u_2,  \mbox{~~~~~~~if}~u_2 \leq 0 ,u_1 \geq 0   \mbox{~and~~}  {ru_2+(1-r)u_1} \geq 0,\\
			\alpha u_1, \mbox{~~~~~~~~~if}~u_2 \leq 0 ,u_1 \geq 0  \mbox{~and~~}  {ru_2+(1-r)u_1} < 0,\\
			-(1-\alpha) u_1, ~~ \mbox{if}~~ u_1  < 0, 
		\end{cases}
		\label{ppl1}
	\end{eqnarray}
 where $0 <r < 1$ is a user-defined parameter and ($u_2$, $u_1$) are errors, representing the deviations of $y$ values from the bounds of PI.  
 \begin{figure}[htp]
		\centering 
 {\includegraphics[width=0.50\linewidth, height = 0.30\linewidth ]{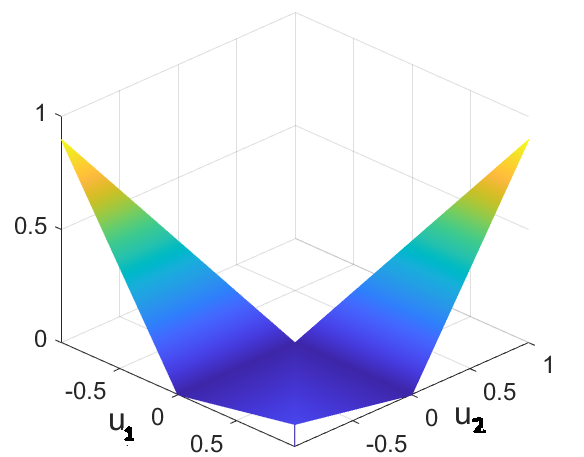}}   
  \caption{Tube loss function for $1-\alpha$ = 0.9.}
  \label{tubeloss}
  \end{figure}
 
 Figure \ref{tubeloss} illustrates the Tube loss for $(1-\alpha) = 0.9$ with $r = 0.5$.  For $r = 0.5$ , the Tube loss function is always a continuous loss function of $u_1$ and $u_2$, symmetrically positioned around the line $u_1 + u_2 = 0$.  In all experiments with a symmetric noise distribution, the \( r \) parameter in the Tube loss function should be set to \( 0.5 \) to capture the denser region of \( y \) values. 

 The Tube loss SVM model seeks a pair of kernel generated functions
 \begin{equation}
     \mu_1(x) = \sum_{i=1}^{m}k(x_i,x)\alpha_i + b_1 \mbox{~~and~}  \mu_2(x) = \sum_{i=1}^{m}k(x_i,x)\beta_i + b_2  \label{11}
 \end{equation}
  by minimizing the optimization problem

  \begin{eqnarray}
\min_{(\alpha, \beta,b_1,b_2)} J{(\alpha, \beta,b_1,b_2)} =  \frac{\lambda}{2}(\alpha^T\alpha + \beta^T\beta) + \sum_{i=1}^{m}\rho_{1-\alpha}^{r} \big (y_i,\big(K(A^T,x_i)\alpha + b_1\big),\big( K(A^T,x_i)\beta + b_2\big)~\big)   \nonumber \\ & \hspace{-180mm} + \delta \sum \limits_{i=1}^{m}  \big|(K(A^T,x_i)(\alpha-\beta) + (b_1-b_2) \big|, \label{prob6}
	\end{eqnarray}
where $\delta, r$ and $\lambda$ are user-defined parameters and $A$ is the $m \times n$ data matrix containing the training set. Further, details on the Tube SVM problem and its minimization using gradient descent method can be found in (\cite{anand2024tube}).  
\subsection{PI estimation through LUBE loss}
The LUBE method (\cite{khosravi2010lower}) was originally developed in the NN framework but, was extended in the SVM framework later in (\cite{shrivastava2014prediction,shrivastava2015prediction}) for probabilistic forecasting of  electric price.  For the given target confidence $(1-\alpha)$, and training set $T$, the LUBE 
 SVM model seeks a pair of kernel generated functions of \ref{11}, $\mu_1(x)$ and $\mu_2(x)$, which are obtained by minimizing the following loss function
 \begin{equation}
		CWC = \frac{1}{R}MPIW \big(1+ \gamma~ (PICP) e^{-\eta(PICP  -(1-\alpha)) } \big). \label{lubecost}
	\end{equation}

 Here, MPIW is the average width of estimated PI on training set, computed by $\frac{1}{m}\sum_{i=1}^{m} (\mu_2(x_i)-\mu_1(x_i))$.  As discussed earlier, the PICP is the coverage of the estimated PI and computed using the  $PICP = \frac{1}{m}\sum_{i=1}^{m}k_i$, where
 \begin{equation}
     k_i = \begin{cases}
			1, ~\mbox{~if~~} y_i \in  [\mu_1(x) , \mu_2(x)]. \\
			0, ~\mbox{Otherwise.}
		\end{cases} \nonumber
 \end{equation}
 Further, the $\gamma(PICP) = \begin{cases}
		0, \mbox{~if ~} PICP \geq 1-\alpha, \\
		1, \mbox{~otherwise,}
	\end{cases}$ , $R$ is the range of response values $y_i$ and $\eta$ is the user-defined parameter. 
	
The major problem with the LUBE cost function (\ref{lubecost}) is that it is very difficult to be optimized because, the PICP is a step function. Khosravi et al. have solved the  LUBE cost function (\ref{lubecost}) using the Particle Swarm Optimization (PSO) (\cite{kennedy1995particle}) to estimate the PI. However, due to sub-optimal solution, high-quality PI is not always observed. Pearce et al. refine the LUBE cost function and use a sigmoidal function to approximate the PICP, allowing the application of the gradient descent method for training  the NN for PI estimating in their work (\cite{pearce2018high}).

\subsection{Comparison of PI estimation SVM models}
In Table \ref{des_prop1}, we  visualize the desirable properties for a PI estimation model and compare the SVM methods in light of  them with a detailed discussion  as follows.

\begin{table}[]
\centering
 \begin{tabular}{|c|c|c|c|c|c|}
 \hline
  &  LS-SVR PI &  SVQR PI &  LUBE  & Tube loss    \\ \hline
   Distribution-free method  & No  & Yes  & Yes  & Yes \\
      Asymptotic guarantees  & \footnotesize {Only with normal noise}  & Yes  & No  &  Yes \\
   Direct PI estimation   &  Yes & No & Yes & Yes  \\
    PI tube movement   & No  & Yes & No & Yes   \\
     Global optimal solution   & Yes  &  Yes & No  & No  \\
      Re-calibration   &  No  & No & Yes  & Yes  \\
       Sparsity   &  No  & No & No  & No  \\
    \hline  
 \end{tabular}
 \caption{ Comparisons of the Quantile , LUBE, QD loss and Tube loss based PI estimation models }
   \label{des_prop1}
 \end{table}

\begin{enumerate}
    \item [(a)] Distribution-free method : - The LS-SVR PI model assumes that the underlying noise distribution of the data is normal and may obtain poor estimate otherwise. In the literature, distribution-based PI methods often struggle to achieve consistent performance across various datasets. The SVQR, LUBE, and Tube loss methods provide PI estimation without assuming any specific distribution, allowing them to generate high-quality PI even in the presence of non-normal noise.
     \item[(b)] Asymptotic guarantees:-  A fundamental requirement in PI estimation models is that the obtained PI should guarantee the target coverage $1-\alpha$ at least asymptomatically, which remain missing in the LUBE method. The LS-SVR PI model provides this guarantee only in presence of normal noise. The SVQR and Tube loss based PI methods provide this asymptotic guarantee.
     \item[(c)] Direct PI estimation:- As detailed in Algorithm 1, the SVQR PI model obtains the two bounds of PI by solving  pair of SVQR problems one by one. The LUBE and Tube loss-based PI models simultaneously obtain the bounds of the PI by solving a single optimization problem.
     \item[(d)] PI tube movement:- The PI tube movement is one another important desirable feature for PI estimation. This movement allows the PI to pass through the denser regions of the data cloud, helping to minimize the width of the PI, while achieving the target coverage. The centered PI is ideal only in the presence of the symmetric noise. However, in the presence of asymmetric noise in the data, the width of the prediction interval (PI) can be reduced by shifting it upward or downward without compromising its coverage.
     The SVQR and Tube loss-based PI models enable PI movement through their parameter $\hat{q}$ and $r$ respectively.

     \item [(e)] Global Optimal Solution:- One of the attractive feature of the standard SVM methods is that they guarantee a global optimal solution by solving a convex optimization problem. However, in PI estimation, only the SVQR and LS-SVR based PI model maintains this guarantee by minimizing the convex loss function in its optimization problems. The Tube loss problem (\ref{prob6}) is non-convex and hence fails to guarantee the global optimal solution. Furthermore, the LUBE problem (\ref{lubecost}) is highly discontinuous and relies on meta-heuristic algorithms for its solution. It  often makes the LUBE solution suboptimal, resulting in poor PI quality.

     \item [(f)] Re-calibration:-  As detailed in Algorithm 1, the SVQR PI model obtains the bound of PI by solving the pair of SVQR problems one-by one. It can not explicitly minimize the width of PI in its optimization problem. This limitation prevents SVQR PI from using the recalibration feature. In recalibration, PI models are retrained to reduce interval width, when  empirical coverage obtained on validation set exceeds the target $1-\alpha$ significantly. During retraining, the PI models increases the value of the parameter ($\delta$ in case of the Tube loss) that trade-off the width of the PI against the coverage in the optimization problem. The LUBE and Tube loss-based models explicitly incorporate the minimization of prediction interval (PI) width in their problems, thereby enabling recalibration, which is practically useful for further reducing the PI width.
     \item [(g)] Sparsity:- Sparsity is yet another promising feature offered by the initial SVM models developed for the classification and regression. However, it remains missing with all PI estimation models developed in the SVM framework.
\end{enumerate}
\section{Sparse  PI estimation in SVM}		
In this section, we introduce the Sparse Support Vector Quantile Regression (SSVQR) model and detail the algorithm for obtaining the sparse PI through it. The SSVQR PI model inherits all properties of SVQR listed in Table \ref{des_prop1} and also brings the sparse solution as well.  Additionally, it addresses the feature selection problem in PI estimation efficiently and also obtain the sparse estimates in CR setting. 

In literature, there are few works that obtain the sparse solution while minimizing the pinball loss function or its variants. Some of the literature such as (\cite{taylor2000quantile, rastogi2018generalized}) obtains the sparse solution for the classification task in the SVM framework while minimizing the variant of the pinball loss function. In (\cite{tanveer2021sparse}), authors have obtained the sparse solution by minimizing a variant of the pinball loss for clustering problem. For quantile estimation, there are a few literature like (\cite{anand2020new, ye2025nonlinear}) which attempt to develop the $\epsilon$-insensitive variant of the pinball loss function, inspired by the $\epsilon$-insensitive loss function used in standard SVM regression (\cite{vapnik2013nature}).

 In view of the above literature, this paper propose a method for obtaining sparse PI estimates in the SVM by formulating the pinball loss minimization problem with $L_1$
 -norm regularization as a linear programming problem (LPP) and further extends  it to address the feature selection problem in PI estimation.

\subsection{Sparse Support Vector Quantile Regression model}

The SSVQR minimizes the pinball loss function with $L_1$- norm regularization.  It seeks the solution of the following problem
\begin{equation}
    \min \limits_{(w,b)} \frac{1}{2} ||w||_1 + C\sum \limits_{i=1}^{m} \rho_q(y_i- (w^T\phi(x_i) +b)), \label{ssvqr1}
\end{equation}
where $C \geq 0$ is the user defined parameter for trading of the regularization against the empirical loss. The solution to problem (\ref{ssvqr1}) is sparse, similar to LASSO regression, as the minimization of the \(L_1\)-norm regularization compels the weight coefficients to shrink to zero.

With the help of the kernel trick,  the representor theorem rewrites the estimated function $ f(x)=(w^T\phi(x) +b) $  as $\sum \limits_{j=1}^{m}k(x_j,x)u_j+ b$. It makes the SSVQR problem (\ref{ssvqr1}) equivalent to
\begin{eqnarray}
     \min \limits_{(u,b,\xi,\xi^*)} \frac{1}{2} ||u||_1 + C\sum \limits_{i=1}^{m} (q\xi_i + (1-q)\xi_i^{*}) \nonumber \\
     & \hspace{-110mm}\mbox{subject to,} \nonumber \\
     & \hspace{-70mm} y_i- \Big( \sum \limits_{j=1}^{m}k(x_j,x_i)u_j+ b \Big) \leq \xi_i, \nonumber  \\
     & \hspace{-70mm} \Big(\sum \limits_{j=1}^{m}k(x_j,x_i)u_j+ b\Big)-y_i \leq \xi_i^{*}, \nonumber \\
     & \hspace{-80mm} \xi_i, \xi_i^{*} \geq 0,~ i =~1,2,..m,
     \label{ssvqr2}
 \end{eqnarray}

Without loss of generality, let us consider the solution vector $u = r-p$, where $r$ and $p$ are vectors of positive numbers i,e., $r_i,p_i > 0 , i=1,2,..,m$ ,then the problem (\ref{ssvqr2}) can be expressed as
\begin{eqnarray}
     \min \limits_{(r, p,b,\xi,\xi^*)} \frac{1}{2} \sum\limits_{i=1}^{m}(r_i + p_i) + C\sum \limits_{i=1}^{m} (q\xi_i + (1-q)\xi_i^{*}) \nonumber \\
     & \hspace{-140mm}\mbox{subject to,} \nonumber \\
     & \hspace{-70mm} y_i- \Big( \sum \limits_{j=1}^{m}k(x_j,x_i)(r_j-p_j)+ b \Big) \leq \xi_i, \nonumber  \\
     & \hspace{-70mm} \Big(\sum \limits_{j=1}^{m}k(x_j,x_i)(r_j-p_j)+ b\Big)-y_i \leq \xi_i^{*}, \nonumber \\
     & \hspace{-80mm} \xi_i, \xi_i^{*},r_i,p_i\geq 0, ~ i =~1,2,..m.
     \label{ssvqr3}
 \end{eqnarray}
The above problem (\ref{ssvqr3}) is a LPP with $4m$ variables, $2m$ linear constraints and $4m$ non-negative constraints, which can be efficiently solved by any LPP solver. The optimal solution  $(r^*,p^*,b^*)$ of the LPP (\ref{ssvqr3})  determines the estimate of the $q^{th}$ quantile function using

\begin{equation}
    f_q(x) = \sum_{i=1}^{m}(r_i^{*}- p_i^{*})k(x_i,x) + b . \label{out2}
\end{equation}
 The asymptotical properties of the SSVQR model remain similar to the SVQR model detailed in (\cite{takeuchi2006nonparametric}).
\subsection{PI estimation through SSVQR}
We detail the algorithm for PI estimation through SSVQR in \ref{alg2}.The SSVQR PI preserves the properties of the SVQR PI, including the global optimal solution, PI tube movement, asymptotic guarantees, and distribution-free estimation, while also achieving a sparse solution.
\begin{algorithm}[h]
\caption{PI estimation through SSVQR}\label{alg:euclid}
\begin{algorithmic}[1]
\Procedure{:- PI through SSVQR} {$T,1-\alpha$}
\State Choose some $\bar{q} \in (0,1-\alpha)$ 
\For{ each $q \in \{\bar{q}, (1+\bar{q}-\alpha)\} $} 
\State Solve the LPP problem (\ref{ssvqr3}) by tuning the value of $C$. Obtain the solution $(r^*,p^*,b^*)$.
\State  Estimate the function $f_q(x)$ using the (\ref{out2}). 
\EndFor\label{euclidendwhile}
\State \textbf{return} $(f_{\bar{q}}(x),f_{1+\bar{q}-\alpha}(x))$
\EndProcedure
\end{algorithmic}
\label{alg2}
\end{algorithm}

\subsection{Feature selection in PI estimation through SSVQR}
Similar to other machine learning tasks, the PI estimation in high-dimensional settings also presents several challenges. The increased dimensionality not only increase the complexity of the PI bounds but also necessitates a larger sample size to ensure the quality of the estimate PI. Therefore, an efficient feature selection method is crucial for reducing the overall complexity of  the PI estimation, particularly when dealing with high-dimensional data.

\begin{algorithm} [h]
\caption{Feature selection through SSVQR}\label{alg:euclid}
\begin{algorithmic}[1]
\Procedure{:- Feature selection through SSVQR} {$T,1-\alpha, \epsilon$}
\State Choose some $\bar{q} \in (0,1-\alpha)$ 
\For{ each $q \in \{\bar{q}, (1+\bar{q}-\alpha)\} $}
\State  Consider the linear kernel  $k(x_i,x_j) = x_i ^{T}x_j$ at the  LPP (\ref{ssvqr3}). 
\State  Solve the LPP (\ref{ssvqr3}) and obtain its solution $(r^*,p^*,b^*)$.
\State Obtain  the $w_q$ using  $ [x_1, x_2,...,x_m ](r^*-p^*)$. 
\EndFor \label{euclidendwhile}
\State  Compute $I_{\bar{q}} = \{ i:,| w_{\bar{q}}(i)| \leq \epsilon \}$ and  $I_{1+\bar{q}-\alpha} = \{ i:,| w_{1+\bar{q}-\alpha}(i) | \leq \epsilon \}$
\State Compute $I = I_{\bar{q}} \cap I_{1+\bar{q}-\alpha} $ and \textit{Feature Set} = $\{1,2,...,m\} - I$
\State \textbf{return}  (\textit{Feature Set})
\EndProcedure
\end{algorithmic}
\end{algorithm}
We detail the feature selection algorithm through SSVQR model for the  linear PI estimation task in Algorithm 3. Here, a linear PI refers to  the PI,  where both bounds are linear functions of the input variables ,i,e. $f_{\bar{q}}(x) = w_{\bar{q}}^{T}x + b_{\bar{q}}$  $f_{1+{\bar{q}}-\alpha}(x) = w_{1+{\bar{q}}-\alpha}^{T}x + b_{1+{\bar{q}}-\alpha}$.

%\textcolor{blue}{Inconsistent symbol $f_{q}(x)$}

The input of Algorithm 3 is the training set $T= \{ (x_i,y_i), x_i \in \mathbf{R}^n, y_i \in \mathbf{R}, i =1,2,....m \}$, specified confidence $(1-\alpha)$, and a very small number $\epsilon$. It finally returns the selected feature set.  In the next section, we  have explored several real-world datasets with numerous features and successfully perform feature selection using the SSVQR method for linear PI estimation, without compromising the quality of the estimated PI.

\subsection{Conformal Regression in SVM}
Finally, we extend the SVM models in CR setting for obtaining the finite sample test set guarantees. In split CR setting, we detail the SVM based CR algorithm as follows.

\begin{algorithm}
\caption{CR through SVQR}\label{alg:euclid1}
\begin{algorithmic}[1]
\Procedure{:- CR through SVQR} {$T,1-\alpha$}
\State Split the training set $T$ into $I_1$ and calibration set $I_2$.
\State Choose some $\bar{q} \in (0,1-\alpha)$ 
\For{ each $q \in \{\bar{q}, (1+\bar{q}-\alpha)\} $} 
\State Solve the QPP problem (\ref{eq2}) on $I_1$  by tuning the value of $C$. Obtain the solution $(\alpha^*,\beta^*)$.
\State  Estimate the function $f_q(x)$ using the (\ref{out1}).
\EndFor\label{euclidendwhile1}

\State  Evaluate the nonconformity score  $E_i  = max \{f_{\bar{q}}(x)-y_i, ~ y_i - f_{1-\bar{q}+\alpha}(x) \} $ on $I_2$. 
\State  Compute the $Q_{1-\alpha}(E,I_2) =  (1-\alpha)(1+ \frac{1}{|I_2|})${-th empirical quantile of } the non-conformity score $E_i$. 
\State \textbf{return} $  C(x_{m+1}) = [ f_{\bar{q}}(x_{m+1}) - Q_{(1-\alpha)}(E,I_2) , f_{\bar{1-q+\alpha}}(x_{m+1}) + Q_{1-\alpha}(E,I_2) ]$
\EndProcedure
\end{algorithmic}
\end{algorithm}

Compared to neural network (NN)-based CR models, the SVM-based CR model offers not only greater interpretability but also more stable prediction sets. In contrast, NN-based CR models often produce varying prediction set across different training runs, even when trained on the same data splits (\( I_1 \) and \( I_2 \)) and with identical hyperparameter settings. It is because that unlike SVMs, NN models often converge to different local optima during training. We empirically verify these advantages of the SVM based CR model in next section.

%In contrast to NN, there is a limited literature which considers the PI estimation task in the SVM framework. The SVM methods are still popular choice over the NN methods.  

\section{Experimental Results}
In this section, we present the numerical results to analyze the quality of the PI obtained by the different SVM models through a series of experiments on simulated/artificial and real-world  benchmark datasets. We also evaluate the effectiveness of the proposed SSVQR model and assess the quality of the PI estimated using it.  We apply our Algorithm 3 for feature selection in high-dimensional real-world benchmark datasets for the linear PI estimation task. We have also compared the SVM based probabilistic forecasting models with the deep forecasting models on several time-series benchmark datasets.

One of the key strengths of SVM machines is their ability to always obtain the global optimal solution, maintaining their relevance and applicability in modern cutting-edge technology.
As detailed in Table \ref{des_prop1}, PI estimation through SVQR and LS-SVR only obtains the global optimal solution but lacks the sparsity. In contrast, the proposed SSVQR PI can obtain the optimal global solution as well as the sparse solution. In view of this, we find the SVQR and LS-SVR model are qualified enough to be compared with the SSVQR model for the PI estimation task in SVM framework. The Tube and LUBE loss PI estimation methods available in SVM framework do not guarantee the global optimal solution and their solution may vary with the choice of initialization. However, we have considered the Tube loss and an improved version of the LUBE loss, the QD loss function (\cite{pearce2018high}) in deep forecasting models to compare their performance with the SVM based probabilistic forecasting models.

\subsection {Evaluation Criteria and Parameter Tunning }
 Now, we describe in detail the evaluation criteria that will be used for our experiments.  In all of our experiments, our aim is to estimate the PI with a confidence level of $1-\alpha = 0.95$ that requires the estimation of the pair of quantile functions $(f_q(x),f_{q+0.95}(x))$, where $0 \leq {q} \leq 0.05 $. In case of artificial datasets, we know the noise distribution and the true quantile function can be easily computed by the inverse of the cumulative distribution function. Therefore, the quality of the quantile function can be accurately assessed by computing the RMSE between the true and estimated quantile functions. In the absence of information about the noise distribution, the quality of the quantile function can be evaluated using Coverage Probability (CP), which measures the fraction of \( y \) values falling below the estimated quantile function. For a \( q^{th} \) quantile estimate, the CP should be as close as possible to \( q \).  For evaluating the  overall quality of PI estimation, we use PICP and MPIW as assessment criteria. An effective PI method must achieve the target \( 1-\alpha \) calibration while minimizing the PI width, as measured by the MPIW value.  Furthermore, we define the Prediction Interval Coverage Error (PICE) as \( \max(0, (1-\alpha) - PICP) \) to quantify the extent to which the model falls short of the target calibration (1-$\alpha$). For comparing PI estimation models, a natural decision criterion is that the model with the lowest PICE should be considered the best. If all models successfully achieve the target calibration, the one with the minimum MPIW should be deemed the most optimal.

To estimate both bounds of the PI, we utilize the RBF kernel, defined as \( k(x_i, x_j) = e^{-q||x_i - x_j||^2} \). As detailed in Algorithm 1 and Algorithm 2, SVQR and SSVQR require solving the QPP (\ref{eq2}) and LPP (\ref{ssvqr3}) twice to obtain the quantile bounds of the PI respectively. We solve the QPPs of the SVQR PI model and the LPPs of the SSVQR model in MATLAB using the 'quadprog' and 'linprog' packages respectively. The SVQR problem(\ref{eq2}) or  SSVQR (\ref{ssvqr3}) problem requires the supply of the two user defined parameters $C$ and  RBF kernel parameter $q$ for non-linear PI estimation. We have tunned the value of the these parameters using the grid search in the search space $\{2 ^{-8},2^{-7},....,2^7,2^8 \} \times \{2 ^{-8},2^{-7},...,2^7,2^8 \} $. 

\subsection{Artificial Datasets}
First, we generate six distinct artificial datasets. In each dataset, the values of \( x_i \) are randomly sampled from a uniform distribution \( U(-5,5) \), while the corresponding \( y_i \) values are obtained by polluting  the function  
$
(1 - x_i + 2x_i^2) e^{-0.5 x_i^2}
$ 
 with different types of noise as described below.
\begin{eqnarray}
    \mbox{ \textbf{AD 1:-}} ~~y_i = (1 - x_i + 2x_i^2) e^{-0.5 x_i^2} + \xi_i , \mbox{where~~}  \xi_i \sim N(0, 0.6) \nonumber \\
    \mbox{ \textbf{AD 2:-}} ~~y_i = (1 - x_i + 2x_i^2) e^{-0.5 x_i^2} + \xi_i , ~~~~~\mbox{where~~}  \xi_i \sim \chi^2(3) \nonumber \\
    \mbox{ \textbf{AD 3:-}} ~~y_i = (1 - x_i + 2x_i^2) e^{-0.5 x_i^2} + \xi_i , \mbox{where~~}  \xi_i \sim N(0, 0.4) \nonumber \\
    \mbox{ \textbf{AD 4:-}} ~~y_i = (1 - x_i + 2x_i^2) e^{-0.5 x_i^2} + \xi_i , \mbox{where~~}  \xi_i \sim N(0, 0.8) \nonumber \\
    \mbox{ \textbf{AD 5:-}} ~~y_i = (1 - x_i + 2x_i^2) e^{-0.5 x_i^2} + \xi_i , \mbox{where~~}  \xi_i \sim U(-5, 5) \nonumber \\
     \mbox{ \textbf{AD 6:-}} ~~y_i = (1 - x_i + 2x_i^2) e^{-0.5 x_i^2} + \xi_i , \mbox{where~~}  \xi_i \sim U(-4, 4) \nonumber 
\end{eqnarray}
In case of each dataset, $2500$ data points are generated in which $1000$ data points are considered for training and rest of them are considered for testing.

% \usepackage{multirow}
\begin{table}[]
{\fontsize{9}{9}\selectfont \begin{tabular}{|l|l|l|l|l|l|l|l|l|}
\cline{1-9}
                      & ($\bar{q},~1+\bar{q}-\alpha$)              & RMSE(Lw,Up)   & Spar (Lw,Up)     & CP (Lw,Up)     & PICP  & PICE & MPIW   & Time (s) \\ \cline{1-9} 
\multirow{5}{*}{SVQR} & (0.01, 0.96)   & (1.8021, 1.4446) & (0\%, 0\%) & (0.0110, 0.9680) & 0.957 & 0  & 2.9054 & 0.4783   \\  
                      & (0.015, 0.965) & (1.7046, 1.4653) & (0\%, 0\%) & (0.0150, 0.9720) & 0.957 & 0  & 2.8244 & 0.5485   \\  
                      & (0.02, 0.97)   & (1.6519, 1.4799) & (0\%, 0\%) & (0.0180, 0.9720) & 0.954 & 0  & 2.7855 & 0.4712   \\  
                      & (0.025, 0.975) & (1.5857, 1.5016) & (0\%, 0\%) & (0.0220, 0.9760) & 0.954 & 0  & 2.7379 & 0.4928   \\ 
                      & (0.03, 0.98)   & (1.5030, 1.5667) & (0\%, 0\%) & (0.0250, 0.9800) & 0.955 & 0  & 2.7212 & 0.6054   \\  \hline
                       \multirow{5}{*}{SSVQR}   &  (0.01, 0.96) & (1.8060, 1.4418) & (15\%, 18\%) & (0.0110, 0.9680) & 0.957 & 0 & 2.9056 & 0.6157 \\  
       & (0.015, 0.965) & (1.7051, 1.4714) & (15\%, 18\%) & (0.0150,  0.9700) & 0.955 & 0 & 2.8275 & 0.5844 \\ 
       & (0.02, 0.97) & (1.6497, 1.4861) & (15\%, 18\%) & 0.0160, 0.9720 & 0.956 & 0 & 2.7939 & 0.6169 \\ 
       & (0.025, 0.975) & (1.5517, 1.4978) & (15\%, 20\%) & (0.0250, 0.9750) & 0.95 & 0 & 2.6995 & 0.5546 \\ 
       & (0.03, 0.98) & (1.5133, 1.5604) & (16\%, 20\%) & (0.0290, 0.9800) & 0.951 & 0 & 2.7288 & 0.515 \\ \hline
\end{tabular}}
\caption{Performance of the SVQR and SSVQR PI models on AD 1 dataset.}
\label{AD1}
\end{table}
\begin{table}[]
{\fontsize{9}{9}\selectfont \begin{tabular}{|l|l|l|l|l|l|l|l|l|}
\cline{1-9}
        & ($\bar{q},~1+\bar{q}-\alpha$)               & RMSE(Lw,Up)   & Spar (Lw,Up)     & CP (Lw,Up)     & PICP  & PICE & MPIW   & Time (s) \\ \cline{1-9} 
\multirow{5}{*}{SVQR} &
        (0.01, 0.96) & (4.2653, 5.7073) & (0\%, 0\%) & (0.0100, 0.9470) & 0.937 & 0.013 & 8.3755 & 0.805 \\ 
        & (0.015, 0.965) & (4.1993, 6.0698) & (0\%, 0\%) & (0.0210, 0.9520) & 0.931 & 0.019 & 8.6865 & 0.699 \\ 
        & (0.02, 0.97) & (4.1312, 6.5751) & (0\%, 0\%) & (0.0220, 0.9610) & 0.939 & 0.011 & 9.1619 & 0.6595 \\
       & (0.025, 0.975) & (4.0525, 6.8163) & (0\%, 0\%) & (0.0340, 0.9650) & 0.931 & 0.019 & 9.3256 & 0.719 \\ 
       & (0.03, 0.98) & (4.0251, 7.4745) & (0\%, 0\%) & (0.0350, 0.9720) & 0.937 & 0.013 & 9.978 & 0.6809 \\ 
        \hline
       \multirow{5}{*}{SSVQR} &  (0.01, 0.96) & (4.1628, 5.5481) & (20\%, 15\%) & (0.0080, 0.9390) & 0.931 & 0.019 & 8.1036 & 0.502 \\ 
         & (0.015, 0.965) & (4.0206, 5.9335) & (18\%, 18\%) & (0.0100, 0.9520) & 0.942 & 0.008 & 8.346 & 0.5173 \\ 
        & (0.02, 0.97) & (4.0156, 6.3077) & (20\%, 25\%) & (0.0110, 0.9640) & 0.953 & 0 & 8.7801 & 0.5128 \\ 
        & (0.025, 0.975) & (3.9444, 7.1596) & (15\%, 20\%) & (0.0240, 0.9680) & 0.944 & 0.006 & 9.5723 & 0.4694 \\ 
        & (0.03, 0.98) & (3.9309, 7.3834) & (20\%, 20\%) & (0.0290, 0.9740) & 0.945 & 0.005 & 9.8075 & 0.5113 \\ \hline
    \end{tabular}}
    \caption{Performance of the SVQR and SSVQR PI models on AD 2 dataset.}
\label{AD2}
\end{table}

\begin{table}
    {\fontsize{9}{9}\selectfont \begin{tabular}{|l|l|l|l|l|l|l|l|l|}
\cline{1-9}
        & ($\bar{q},~1+\bar{q}-\alpha$)               & RMSE(Lw,Up)   & Spar (Lw,Up)     & CP (Lw,Up)     & PICP  & PICE & MPIW   & Time (s) \\ \cline{1-9} 
\multirow{5}{*}{SVQR} & 
        (0.01, 0.96) & (1.7894, 1.5837) & (0\%, 0\%) &( 0.0120, 0.9570) & 0.945 & 0 & 2.9 & 0.4872 \\ 
       & (0.015, 0.965) & (1.7355, 1.6297) & (0\%, 0\%) & (0.0130, 0.9590) & 0.946 & 0.004 & 2.887 & 0.5077 \\ 
        & (0.02, 0.97) & (1.5696, 1.6407) & (0\%, 0\%) & (0.0230, 0.9590) & 0.936 & 0.014 & 2.7223 & 0.5004 \\ 
      &  (0.025, 0.975) & (1.5070, 1.6927) & (0\%, 0\%) & (0.0280, 0.9680) & 0.94 & 0.01 & 2.71 & 0.4907 \\ 
       & (0.03, 0.98) & (1.4694, 1.7477) & (0\%, 0\%) & (0.0300, 0.9750) & 0.945 & 0.005 & 2.7394 & 0.4745 \\ 
        \hline
   \multirow{5}{*}{SSVQR} &      (0.01, 0.96) & 1.8411, 1.5781 & 20\%, 40\% & 0.0120, 0.9570 & 0.945 & 0.005 & 2.9462 & 0.4367 \\ 
         & (0.015, 0.965) & (1.8297, 1.5917) & (40\%, 30\%)& (0.0120, 0.9570) & 0.945 & 0.005 & 2.9458 & 0.4563 \\ 
        & (0.02, 0.97) & (1.6999, 1.6969) & (40\%, 40\%) & (0.0180, 0.9600) & 0.942 & 0.008 & 2.9026 & 0.4671 \\ 
       & (0.025, 0.975) & (1.6509, 1.7191) & (40\%, 40\%) & (0.0220, 0.9620) & 0.94 & 0.01 & 2.8817 & 0.4501 \\ 
      &  (0.03, 0.98) & (1.4865, 1.7721) & (30\% , 40\%) & (0.0290, 0.9700) & 0.941 & 0.009 & 2.7739 & 0.4717 \\ \hline
    \end{tabular}}
      \caption{Performance of the SVQR and SSVQR PI models on AD 3 dataset.}
\label{AD3}
\end{table}
\begin{table}
    {\fontsize{9}{9}\selectfont \begin{tabular}{|l|l|l|l|l|l|l|l|l|}
\cline{1-9}
        & ($\bar{q},~1+\bar{q}-\alpha$)               & RMSE(Lw,Up)   & Spar (Lw,Up)     & CP (Lw,Up)     & PICP  & PICE & MPIW   & Time (s) \\ \cline{1-9} 
\multirow{5}{*}{SVQR} & 
        (0.01, 0.96) & (2.5953, 2.1197) & (0\%, 0\% )& (0.0160, 0.9530) & 0.937 & 0.013 & 4.1549 & 0.723 \\ 
         & (0.015, 0.965) & (2.5595, 2.1292) & (0\%, 0\%) & (0.0160, 0.9530) & 0.937 & 0.013 & 4.1268 & 0.6713 \\ 
       & (0.02, 0.97) & (2.4221, 2.2193) & (0\%, 0\%) & (0.0200, 0.9610) & 0.941 & 0.009 & 4.0817 & 0.6704 \\ 
       & (0.025, 0.975) & (2.2561, 2.2555) & (0\%, 0\%) & (0.0290, 0.9630) & 0.934 & 0.016 & 3.9365 & 0.6815 \\ 
       & (0.03, 0.98) & (2.2211, 2.3699) & (0\%, 0\%) & (0.0310, 0.9670) & 0.936 & 0.014 & 4.0273 & 0.6343 \\ \hline
    \multirow{5}{*}{SSVQR}   &  (0.01, 0.96) & (2.5842, 2.1364) & (20\%, 40\%) & (0.0160, 0.9530) & 0.937 & 0.013 & 4.1577 & 0.4704 \\
       & (0.015, 0.965) & (2.5475, 2.1911) & (20\%, 20\%) & (0.0160, 0.9560) & 0.94 & 0.01 & 4.1802 & 0.4745 \\ 
       & (0.02, 0.97) & (2.4939, 2.2455) & (20\%, 20\%) & (0.0190, 0.9610) & 0.942 & 0.008 & 4.183 & 0.4818 \\ 
       & (0.025, 0.975) & (2.3617, 2.3037) & (40\%, 20\%) & (0.0240, 0.9640) & 0.94 & 0.01 & 4.1002 & 0.4662 \\ 
       & (0.03, 0.98) & (2.2689, 2.4147) & (40\%, 20\%) & (0.0290, 0.9700) & 0.941 & 0.009 & 4.121 & 0.4884 \\ \hline
    \end{tabular}}
   \caption{Performance of the SVQR and SSVQR PI models on AD 4 dataset.}
\label{AD4}
\end{table}

\begin{table}
    {\fontsize{9}{9}\selectfont \begin{tabular}{|l|l|l|l|l|l|l|l|l|}
\cline{1-9}
        & ($\bar{q},~1+\bar{q}-\alpha$)               & RMSE(Lw,Up)   & Spar (Lw,Up)     & CP (Lw,Up)     & PICP  & PICE & MPIW   & Time (s) \\ \cline{1-9} 
\multirow{5}{*}{SVQR} & 
        (0.01, 0.96) & (4.9694, 4.4016) & (0\%, 0\% )& (0.0100, 0.9560) & 0.946 & 0.004 & 8.1044 & 0.7917 \\
     &   (0.015, 0.965) & (4.8456, 4.4539) & (0\%, 0\%) & (0.0180, 0.9600) & 0.942 & 0.008 & 8.0257 & 0.8009 \\ 
      &  (0.02, 0.97) & (4.8100, 4.4860) & (0\%, 0\%) & (0.0190, 0.9620) & 0.943 & 0.007 & 8.0223 & 0.736 \\ 
       & (0.025, 0.975) & (4.7111, 4.5143) & (0\%, 0\%) & (0.0280, 0.9640) & 0.936 & 0.014 & 7.943 & 0.6895 \\
      &  (0.03, 0.98) & (4.6921, 4.5368) & (0\%, 0\%) & (0.0290, 0.9680) & 0.939 & 0.011 & 7.9481 & 0.6191 \\ \hline
       \multirow{5}{*}{SSVQR}      & (0.01, 0.96) & (5.2658, 4.5044) & (40\%, 30\%) & (0.0050, 0.9610) & 0.956 & 0 & 8.5344 & 0.5394 \\ 
        & (0.015, 0.965) & (4.8787, 4.5073) & (35\%, 30\%) & (0.0160, 0.9610) & 0.945 & 0.005 & 8.1171 & 0.4835 \\ 
        & (0.02, 0.97) & (4.8138, 4.5249) & (30\%, 30\%) & (0.0190, 0.9650) & 0.946 & 0.004 & 8.0659 & 0.472 \\ 
       & (0.025, 0.975) & (4.7464, 4.6037) & (25\%, 30\%) & (0.0260, 0.9670) & 0.941 & 0.009 & 8.0692 & 0.5102 \\ 
        & (0.03, 0.98) & (4.6962, 4.7637) & (30\%, 40\%) & (0.0290, 0.9720) & 0.943 & 0.007 & 8.1832 & 0.5068 \\ \hline
    \end{tabular}}
   \caption{Performance of the SVQR and SSVQR PI models on AD 5 dataset.}
\label{AD5}
\end{table}

\begin{table}
    {\fontsize{9}{9}\selectfont 
    \begin{tabular}{|l|l|l|l|l|l|l|l|l|}
\cline{1-9}
        & ($\bar{q},~1+\bar{q}-\alpha$)              & RMSE(Lw,Up)   & Spar (Lw,Up)     & CP (Lw,Up)     & PICP  & PICE & MPIW   & Time (s) \\ \cline{1-9} 
\multirow{5}{*}{SVQR} & 
        (0.01, 0.96) & (6.1510, 5.5057) & (0\%, 0\%) & (0.0110, 0.9630) & 0.952 & 0 & 10.0826 & 0.6437 \\ 
      &  (0.015, 0.965) & (6.0517, 5.5584) & (0\%, 0\%) & (0.0150, 0.9670) & 0.952 & 0 & 10.031 & 0.6798 \\ 
     &   (0.02, 0.97) & (5.9777, 5.6130) & (0\%, 0\% )& (0.0170, 0.9700) & 0.953 & 0 & 10.0141 & 1.0752 \\ 
     &   (0.025, 0.975) & (5.8879, 5.6868) & (0\%, 0\%) & (0.0230, 0.9760) & 0.953 & 0 & 9.9987 & 0.9265 \\ 
     &   (0.03, 0.98) & (5.7599, 5.7156) & (0\%, 0\%) & (0.0290, 0.9780) & 0.949 & 0.001 & 9.8878 & 0.9445 \\ \hline
        
       \multirow{5}{*}{SSVQR} &  (0.01, 0.96) & (5.9304, 5.4581) & (10\%, 15\%) & (0.0110, 0.9610) & 0.95 & 0 & 9.7944 & 0.569 \\ 
      &  (0.015, 0.965) & (5.9107, 5.5686) & (10\%, 20\%) & (0.0140, 0.9700) & 0.956 & 0 & 9.9016 & 0.5656 \\ 
      &  (0.02, 0.97) & (5.7836, 5.5880) & (15\%, 25\%) & (0.0180, 0.9700) & 0.952 & 0 & 9.7807 & 0.5766 \\ 
      &  (0.025, 0.975) & (5.6486, 5.6038) & (20\%, 20\%) & (0.0260, 0.9710) & 0.945 & 0.005 & 9.6443 & 0.5423 \\ 
      &  (0.03, 0.98) & (5.6184, 5.6506) & (10\%, 15\%) & (0.0300, 0.9760) & 0.946 & 0.004 & 9.6653 & 0.5636 \\ \hline
    \end{tabular}}
    \caption{Performance of the SVQR and SSVQR PI models on AD 6 dataset.}
\label{AD6}
\end{table}
\subsection{ Artificial Datasets Results, Discussion and Analysis}
\begin{figure}
     \centering
     \subfloat[]{\includegraphics[width=0.50\linewidth, height = 0.3\linewidth]{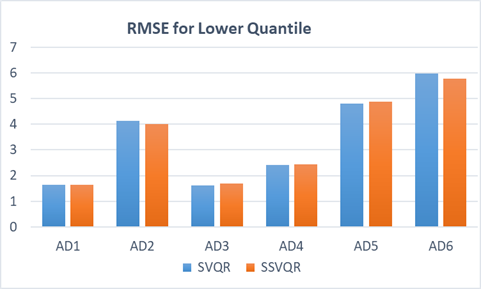}\label{<figure1>}}~~
     \subfloat[]{\includegraphics[width=0.50\linewidth, height = 0.3\linewidth]{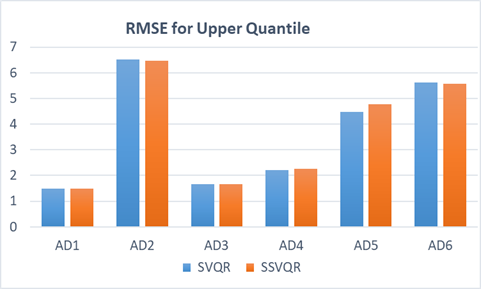}\label{<figure2>}}
     \caption{Comparison of the quality of quantile function obtained by the SVQR and SSVQR PI models.}
     \label{steady_state}
\end{figure}

We present the performance of the SVQR and SSVQR model for PI estimation task with the different values of $\bar{q}$ for each of simulated dataset in Table \ref{AD1}-\ref{AD6}.  The rightmost column of these Tables list the different pairs of target quantiles ($\bar{q},~1+\bar{q}-\alpha$), required for the PI estimation with $0.95$ confidence level. As detailed in Section 5.1 of this paper, for artificial datasets, the quality of the estimated upper and lower quantiles of the PI can be best evaluated by computing the RMSE between the estimated quantiles and their corresponding true quantile functions. The third column of the Tables \ref{AD1}-\ref{AD6} list these RMSE for different values of $\bar{q}$ and mean of them are plotted at Figure \ref{steady_state} for lower and upper quantile estimation separately for different simulated datasets. The quantile bounds estimated by the SSVQR model are comparable to, or slightly better than, those obtained from the SVQR model. Replacing $L_2$ regularization with $L_1$ regularization in the SVM quantile regression model does not result in significantly different estimates. However, the major advantage of using the SSVQR model is its ability to obtain the sparse solution vector. Figure \ref{fig:enter-label}(a) plots the sparsity of the solution vector corresponding to the upper and lower quantile bounds obtained by the SSVQR model.  It highlights that the SSVQR model effectively reduces significant coefficients of the solution vector to near zero which enables the feature selection task in PI estimation and also simplify the overall prediction process.

\begin{figure}
    \centering
   \subfloat[]{\includegraphics[width=0.52\linewidth,height = 0.3\linewidth]{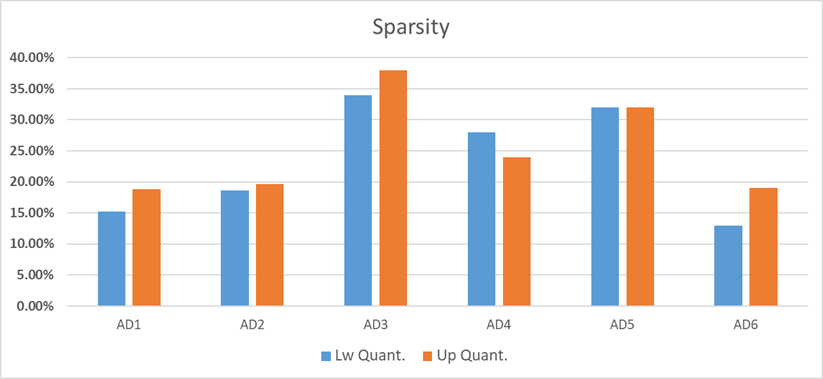}}
   \subfloat[]{\includegraphics[width=0.52\linewidth,height = 0.3\linewidth]{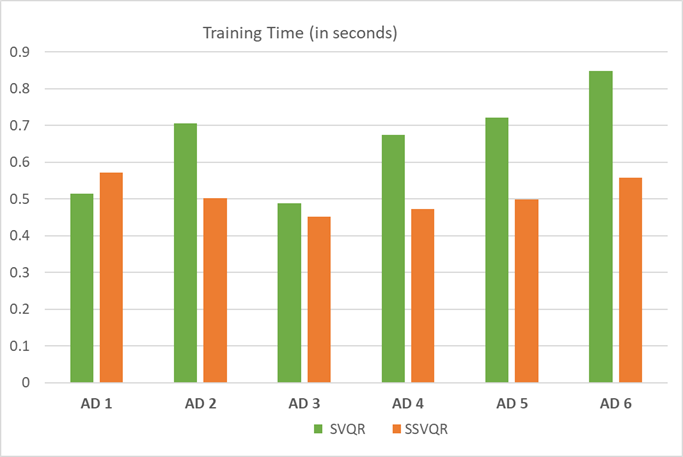}}
    \caption{(a) Qunatile functions estimated by the proposed SSVQR model is sparse while SVQR model fails to obtain the sparse solution. (b) Average training time comparison of the SVQR and SSVQR models for PI estimation.}
    \label{fig:enter-label}
\end{figure}

We compare the overall average training time (in seconds) taken by the SVQR PI and SSVQR PI models for the PI estimation task across different artificial datasets in Figure \ref{fig:enter-label}(b).  It shows that the SSVQR requires fewer seconds train the PI model than the SVQR model.  We have solved the QPPs of the SVQR PI model and LPPs of the SSVQR model in MATLAB with 'quadprog' and 'linprog' packages respectively.

\begin{figure}
     \centering
     \subfloat[]{\includegraphics[width=0.58\linewidth]{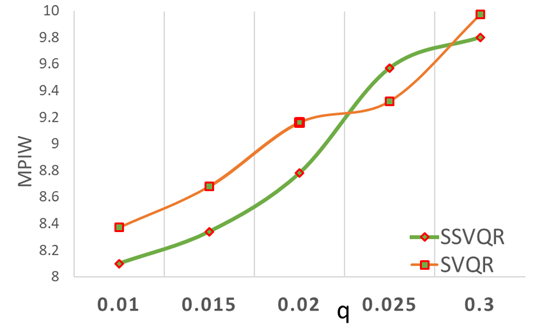}}
     \subfloat[]{\includegraphics[width=0.51\linewidth]{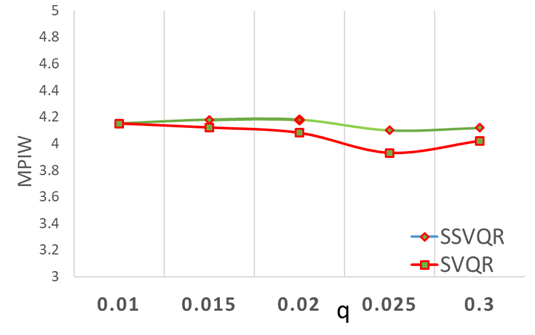}}
     \caption{Plot of the MPIW obtained by the SVQR and SSVQR PI models against $\bar{q}$ on  (a) AD2 and (b) AD4 dataset.}
     \label{steady_state11}
\end{figure}

Another key observation from the numerical results in Tables \ref{AD1}-\ref{AD6} is the consistent performance of the SVQR and SSVQR PI models across different datasets. In all cases, both the SVQR and SSVQR PI models manage to approximately achieve $95\%$ target coverage. We plot the MPIW values obtained by the SSVQR and SVQR PI models against different values of $\bar{q}$ on dataset AD2 and AD4 in  Figure \ref{steady_state11}. On AD2 dataset, the MPIW values of the PI obtained by both models increase with $\bar{q}$. It is  evident from Table \ref{AD2} that this increase is not related to the PICP values obtained by the SVQR and SSVQR PI models. Actually, it is caused by the nature of noise present in the AD2 dataset.  The AD2 dataset contains asymmetric noise from the ($\chi^{2}$) distribution, leading to a higher density of data points in the lower region of the input-target space, which gradually decreases as we move upward.
As $\bar{q}$ decreases, the resultant PI shifts downward, passing through denser regions of the data cloud, leading to lower MPIW values. It shows that the movement of the PI tube due to change of $\bar{q}$ values may lead to the narrower PI particularly in presence of the asymmetric noise in the data. Apart from the AD2 dataset, all other artificial datasets contain noise from symmetric distributions. In these datasets, the centered PI ($(f_{0.025}(x), f_{0.975}(x))$ is expected to achieve the high quality PI. Figure \ref{steady_state11}(b) shows that at $q = 0.025$, both the SVQR and SSVQR PI models attain the lowest MPIW values on AD4 dataset.
  
\subsection{Feature Selection through SSVQR}
Next, we apply our SSVQR model for feature selection in PI estimation with linear kernel. To demonstrate this, we use five popular real-world benchmark datasets 
namely Spambase, Student Performance, Boston Housing, UCI-secom and MADELON. We use the $80\%$ of data points for training and use rest of them for testing. We set the target calibration $1-\alpha=$ 0.95.  Following Algorithm 3, we perform feature selection using the SSVQR PI model for  $\bar{q} =0.025$ and present the results in Table \ref{tab23}.
The numerical results clearly demonstrate that the SSVQR PI feature selection (as detailed in Algorithm 3) can significantly reduce the number of features while maintaining the quality of the PI, as measured by PICP and MPIW. On average, it could reduce the $69\%$ of features on considered  datasets and reduce the complexity of the PI estimation task significantly. The training time of the PI estimating task improves significantly after dropping irrelevant features through Algorithm 3. On average, it could improve the $53\%$ improvement in training time. Also,  reducing the significant numbers of features will reduce the tunning and testing time of the  PI model.

For high-dimensional datasets such as UCI secom and MADELON, feature selection leads to a significant improvement in the quality of the estimated PI in terms of its coverage. The SSVQR  based feature selection algorithm eliminates a significant number of irrelevant features and learn the PI in  relatively much lower dimension which results in better PI coverage with linear functions. Table \ref{tab:dropped_features} lists the features dropped by the SSVQR-PI feature selection algorithm for each dataset.

%, constructed by the pair of linear functions.  This improvement arises because, as dimensionality increases, the complexity of the underlying function required to accurately capture the quantile functions for PI estimation can grow exponentially.

%By eliminating a , the complexity of the PI estimation task is reduced, enabling high-quality PIs to be estimated even using a set of linear functions. 
\begin{table}[ht]
\centering
\renewcommand{\arraystretch}{1.5} % Increase row height
\scriptsize
\begin{tabular}{|c|c|ccc|ccc|c|}
\hline
\textbf{Dataset} & \textbf{Dimension} & \multicolumn{3}{c|}{\textbf{Before Feature Selection}} & \multicolumn{3}{c|}{\textbf{After Feature Selection}} & \textbf{\% Reduced} \\
\cline{3-9}
 &  & \textbf{PICP} & \textbf{MPIW} & \textbf{ Train Time(s)} & \textbf{PICP} & \textbf{MPIW} & \textbf{ Train Time(s)} & \textbf{Features} \\
\hline
Spambase & (4601, 58) & 0.9663 & 0.9330 & 60  & 0.9653 & 0.9340 & 52 & 46\% \\
\hline
Student Perf. & (395, 16) & 0.8861 & 6.8429 & 0.48  & 0.8861 & 6.8429 & 0.42  & 73\% \\
\hline
Boston Housing & (505, 14) & 0.9406 & 23.0647 & 0.81  & 0.9406 & 23.0647 & 0.7 & 38\% \\
\hline
uci-secom & (1568, 591) & 0.8758 & 1.6146 & 16  & 0.9204 & 1.6683 & 7 & 91\% \\
\hline
MADELON & (2000, 500) & 0.7100 & 2.0007 & 58  & 0.9375 & 2.0020 & 11  & 99\% \\
\hline
\end{tabular}
\caption{Performance comparison before and after feature selection using the SSVQR PI model}
\label{tab23}
\end{table}

\begin{table}[htbp]
\centering
\scriptsize
\begin{tabular}{|l|p{14cm}|}
\hline
\textbf{Dataset} & \textbf{Dropped Features} \\ \hline
Spambase & 2, 8, 11, 13, 16, 17, 18, 19, 20, 21, 22, 23, 25, 27, 31, 35, 37, 46, 47, 48, 49, 50, 53, 54, 55, 56 \\ \hline
Student Perf. & 1, 2, 3, 4, 5, 6, 7, 8, 9, 10, 11 \\ \hline
Boston Housing & 2, 3, 4, 7, 10 \\ \hline
UCI Secom  & 1, 4-20, 22-27, 28-41, 42-50, 52-58, 60-66, 68-70, 72, 74-87, 89, 91-110, 112-132, 134, 137-138, 140-157, 159-160, 162-187, 189-203, 205-224, 226-246, 247-296, 297-332, 333-362, 364-386, 387-400, 401-418, 420-422, 424-431, 434-435, 437-438, 440-466, 467, 469-481, 483, 489-498, 501-509, 512-520, 522-538, 540-545, 546, 548-560, 563-569, 571, 573-579, 580, 582-587 \\ \hline
MADELON & 0-89, 91-227, 229-275, 277-331, 333-444, 446-499 \\ \hline
\end{tabular}
\caption{Feature Selection by SSVQR PI}
\label{tab:dropped_features}
\end{table}

\subsection{ SVM PI estimation methods on benchmark datasets}
We have done experiments on the two popular benchmark datasets namely Boston Housing and Concrete  and evaluate the quality of the PI estimated by the SSVQR, SVQR and LS-SVR based PI estimation methods for different value of the $\bar{q}$ with non-linear RBF kernel.   Table \ref{BS} and \ref{tab_para} contains the numerical results obtained on the  Boston Housing and Concrete datasets respectively.  We can observe that the SSVQR, SVQR and LS-SVR PI models obtains a similar quality of the PI but SSVQR models always obtain the sparse solution vector. 
\begin{table}[h]
    \centering
    {\fontsize{9}{10} \selectfont
    \begin{tabular}{|c|c|c|c|c|c|c|c|c|}
    \hline
      &  q & Spar (Lw, Up) & CP (Lw, Up) & PICP & PICE & MPIW & Time (s) \\ \hline
    \multirow{3}{*}{SVQR}   & (0.025, 0.925) & (0\%, 0\%) & (0.028, 0.930) & 0.90 & 0 & 28.55 & 0.1347 \\
       & (0.05, 0.95) & (0\%, 0\%) & (0.052, 0.950) & 0.90 & 0.00 & 33.06 & 0.1472 \\
     &   (0.075, 0.975) & (0\%, 0\%) & (0.088, 0.972) & 0.88 & 0.02 & 38.23 & 0.1670 \\  \hline
     \multirow{3}{*}{SSVQR}  & (0.025, 0.925) & (17\%, 16\%) & (0.027, 0.927) & 0.90 & 0  & 28.62 & 0.0657 \\ 
      &  (0.05, 0.95) & (15\%, 22\%) & (0.048, 0.947) & 0.90 & 0 & 32.77 & 0.0678 \\ 
     &   (0.075, 0.975) & (14\%, 28\%) & (0.080, 0.967) & 0.89 & 0.01 & 38.24 & 0.0681 \\ \hline
      \multirow{3}{*}{LS-SVR} &  (0.025, 0.925) & (0\%,0\% )&  & 0.91 & 0 & 31.19 & 0.0064 \\ 
     &(0.050, 0.950) & (0\%,0\%) &  & 0.91 & 0 & 30.18 & 0.0067 \\
        
    & (0.075, 0.975) & (0\%,0\%) &  &0.89 & 0.01 & 31.19 & 0.0060 \\ \hline
    \end{tabular}}
    \caption{Comparison of different SVM PI estimation methods on Boston Housing}
    \label{BS}
\end{table}
\begin{table}[h]
    \centering
    {\fontsize{9}{10} \selectfont
    \begin{tabular}{|c|c|c|c|c|c|c|c|c|}
    \hline
      &  q & Spar (Lw, Up) & CP (Lw, Up) & PICP & PICE & MPIW & Time (s) \\ \hline
    \multirow{3}{*}{SVQR}  & (0.025, 0.925) & (0\%, 0\%) & (0.0219, 0.8997) & 0.8777 & 0.0723 & 32.5492 & 2.6141 \\ 
       & (0.05, 0.95) & (0\%, 0\%) & (0.0408, 0.9436) & 0.9028 & 0.0472 & 28.6541 & 2.8405 \\ 
       & (0.075, 0.975) & (0\%, 0\%) & (0.0690, 0.9561) & 0.8871 & 0.0629 & 32.1635 & 2.5197 \\ \hline
  \multirow{3}{*}{SSVQR}   &  (0.025, 0.925) & (12\%, 12\%) & (0.0340, 0.9029) & 0.8689 & 0.0811 & 29.4962 & 0.1858 \\ 
      &  (0.05, 0.95) & (12\%, 12\%) & (0.0583, 0.9272) & 0.8689 & 0.0811 & 27.8917 & 0.1795 \\ 
      &  (0.075, 0.975) & 12\%, 12\% & (0.0777, 0.9515) & 0.8738 & 0.0762 & 30.2547 & 0.1837 \\ \hline
    \multirow{3}{*}{LS-SVR}   &    (0.025, 0.925) &0\%,0\% & & 0.9126 & 0.0374 & 28.2429 & 0.4125 \\ 
     &   (0.050, 0.950) &(0\%,0\%)& & 0.8932 & 0.0568 & 27.3308 & 0.4631 \\
      &  (0.075, 0.975) &(0\%,0\%)&  &0.9029 & 0.0471 & 28.2429 & 0.3954 \\ \hline

    \end{tabular}}
    \caption{Comparison of different SVM PI estimation methods on Concrete dataset}
    \label{tab_para}
\end{table}

\subsection{ Probabilistic Forecasting  with SVM models}
In this section, we compare the performance of the  proposed SSVQR, SVQR and LS-SVR model for the probabilistic forecasting. We also train several recent and widely adopted deep learning models for probabilistic forecasting developed in a distribution-free setting, including Quantile-based LSTM, Tube Loss LSTM, and Quality-Driven (QD) Loss LSTM models \cite{pearce2018high} to compare their performance against the SVM-based models. The QD loss  \cite{pearce2018high} is the improved version of the LUBE model which can be used minimized with the gradient descent method in deep learning architecture. Also, we have added the popular Deep AR (\cite{salinas2020deepar}) probabilistic forecasting method for baseline comparison.

We consider three popular time-series datasets namely Female Births (365 $\times$ 1), Minimum Temperature (3651 $\times$ 1)  and Beer Production (464 $\times$ 1). We have used the 70$\%$ of dataset as training set and rest of them are used for testing. Out of the training set, the last $10\%$ of the observations have been used for the validation set. We present the numerical results in Table \ref{pfr}.  All of the models were asked to obtain the probabilistic  forecast for target calibration $1-\alpha = 0.95$. 

\begin{figure}
    \centering
     \subfloat[]{\includegraphics[width=1.0\textwidth, height = 0.25\textwidth]{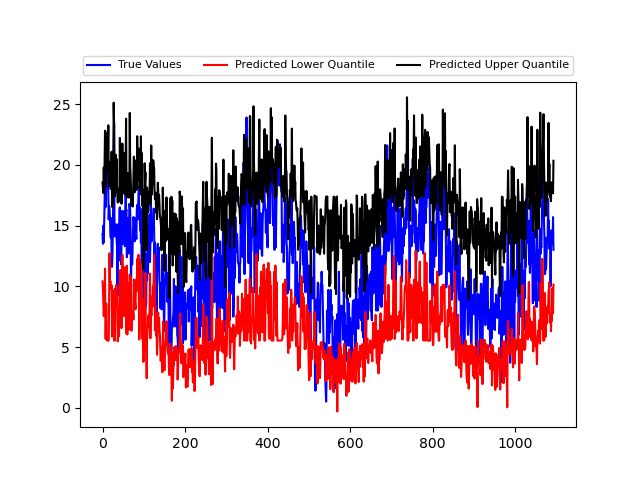}} \\
     \subfloat[]{ \includegraphics[width=1.0\textwidth, height = 0.25\textwidth]{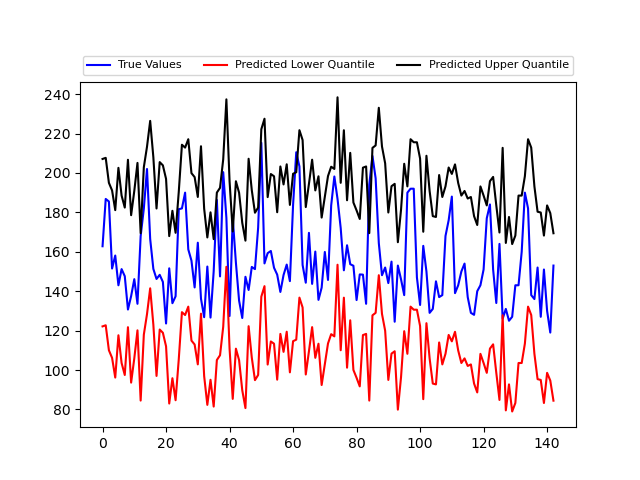}}
    \caption{Probabilistic forecasting with SSVQR model on daily 
 (a) Temperature and (b)  Beer Production dataset}
    \label{fig:daily_birth}
\end{figure}

One major observation from the Table \ref{pfr} is that the SVM based probabilistic forecasting models obtain competitive performance compared to complex LSTM based deep forecasting models after efficient tuning of its parameters. Also, the SSVQR based probabilistic forecasting model obtains the sparse solution.  Table \ref{tab_para1} presents the tuned neural architectures of the LSTM models, along with the number of weights to be optimized for each of the probabilistic forecasting models used.
The LSTM based probabilistic forecasting models are more complex and requires the optimization of thousands of weights where as the SVQR based probabilistic models are much simpler architectures and could obtain similar quality of the PI as obtained by the LSTM based models.  Figure (\ref{fig:daily_birth}) (a) and (b) shows the forecasting of the SSVQR and SVQR model on Temperature and Beer Production datasets respectively.

\begin{table}[http]
\centering
{\fontsize{9}{9}\selectfont %
\begin{tabular}{|l| l| c| c| c| c|}
\hline
\textbf{Dataset} & \textbf{Method} & \textbf{PICP} & \textbf{MPIW} & \textbf{Training Time} & \textbf{Sparsity} \\ \hline
\multirow{7}{*}{\textbf{Female Births}}
 & SSVQR         & 0.93  & 28.00    & 1.12    & 61 \% \\
 & SVQR          & 0.95  & 27.11    & 0.97    & 0 \% \\
 & LS-SVR        & 0.95  & 37.70    & 3.60    & 0 \%  \\ \cline{2-6}
 & Quantile LSTM & 0.95  & 28.20    & 118.00  & -- \\
 & Tube LSTM     & 0.96  & 28.09    & 43.00   & -- \\
 & QD LSTM       & 0.94  & 38.98    & --      & -- \\ 
 & DeepAR   & 0.94 & 29.8 & 55.0 & 0 \% \\ \hline

\multirow{7}{*}{\textbf{Minimum Temp.}}
 & SSVQR         & 0.96  & 10.72    & 200.79  & 69 \% \\
 & SVQR          & 0.96  & 75.81    & 172.00  & 0 \% \\
 & LS-SVR        & 0.95  & 10.59    & 0.53    & 0 \% \\ \cline{2-6}
 & Quantile LSTM & 0.95  & 24.82    & 1135.00 & -- \\
 & Tube LSTM     & 0.94  & 15.56    & 447.00  & -- \\
 & QD LSTM       & 0.79  & 5.94     & --      & -- \\ 
 & DeepAR    & 0.90 & 12.79 & 58.0 & 0 \% \\ \hline

\multirow{7}{*}{\textbf{Beer Prod.}}
 & SSVQR         & 0.96  & 0.94     & 1.77    & 70 \% \\
 & SVQR          & 0.95  & 75.73    & 0.78    & 0 \% \\
 & LS-SVR        & 0.96  & 79.48    & 0.29    & 0 \%  \\ \cline{2-6}
 & Quantile LSTM & 0.94  & 134.80   & 132.80  & -- \\
 & Tube LSTM     & 0.95  & 42.91    & 89.60   & -- \\
 & QD LSTM       & 0.96  & 159.71   & --      & -- \\ 
 & DeepAR    & 0.76 & 12.43 & 62.0 & 0 \% \\ \hline
\end{tabular}%
}
\caption{Performance comparison of SVM and deep learning based probabilistic forecasting methods on benchmark datasets.}
\label{pfr}
\end{table}
\begin{table}[h]
\centering
{\fontsize{9}{9}\selectfont
\begin{tabular}{|l|l|l|l|l|}
\hline
\textbf{Dataset} & \textbf{Model} & \textbf{Architecture} & \textbf{Weights} & \textbf{W.\ Size} \\ \hline
\multirow{9}{*}{\textbf{Female Births}}
 & SVQR          & Kernel           & 256   & 10  \\
 & SSVQR         & Kernel           & 256   & 15  \\
 & LS-SVR        & Kernel           & 256   & 20  \\ \cline{2-5}
 & Quantile LSTM & LSTM [100]       & 30 K  & 25  \\
 & Tube LSTM     & LSTM [100]       & 30 K  & 25  \\
 & QD LSTM       & LSTM [100]       & 30 K  & 25  \\ 
 & DeepAR   & LSTM [40,40]     & 13 K  & 28  \\ \hline

\multirow{9}{*}{\textbf{Minimum Temp.}}
 & SVQR          & Kernel           & 2 556 & 12  \\
 & SSVQR         & Kernel           & 2 556 & 18  \\
 & LS-SVR        & Kernel           & 2 556 & 22  \\ \cline{2-5}
 & Quantile LSTM & LSTM [16,8]      & 32 K  & 28  \\
 & Tube LSTM     & LSTM [16,8]      & 32 K  & 28  \\
 & QD LSTM       & LSTM [16,8]      & 32 K  & 28  \\ 
 & DeepAR    & LSTM [40,40]     & 13 K  & 56  \\ \hline

\multirow{9}{*}{\textbf{Beer Prod.}}
 & SVQR          & Kernel           & 325   & 8   \\
 & SSVQR         & Kernel           & 325   & 14  \\
 & LS-SVR        & Kernel           & 325   & 18  \\ \cline{2-5}
 & Quantile LSTM & LSTM [64,32]     & 29 K  & 24  \\
 & Tube LSTM     & LSTM [64,32]     & 29 K  & 24  \\
 & QD LSTM       & LSTM [64,32]     & 29 K  & 24  \\ 
 & DeepAR    & LSTM [40,40]     & 13 K  & 48  \\ \hline
\end{tabular}}%
\caption{ Comparison of the complexity of  used SVM and deep learning based probabilistic forecasting models on benchmark datasets. LSTM  [16,8] means that the used LSTM model has two hidden layer containing 16 and 8 neurons respectively. }
\label{tab_para1}
\end{table}

%\begin{table}[htbp]
%\centering
%\caption{Other Parameters for First Four Methods}
%\label{tab:performance_parameters}
%\tiny
%\resizebox{\textwidth}{!}{%
%\begin{tabular}{|l| l |c| c| c| c|}
%\hline
%\textbf{Dataset} & \textbf{Method} & \textbf{Window Size} & \textbf{S} & \textbf{c3} & \textbf{c1} \\ \hline
%\multicolumn{6}{l}{\textbf{Female Births}} \\ \hline
% & LP-SVQR      & 3    & 0.01   & 32  & 5   \\ 
% & ESVQR        & 5    & 19.00  & --  & 15  \\ 
% & LS-SVR       & 3    & 0.02   & --  & 128 \\ \hline
%\multicolumn{6}{l}{\textbf{Minimum Temperature}} \\ \hline
% & LP-SVQR      & 2    & 0.01   & 64  & 8   \\ 
% & ESVQR        & 12   & 16.00  & --  & 10  \\ 
% & LS-SVR       & 12   & 0.03   & --  & 128 \\ \hline
%\multicolumn{6}{l}{\textbf{Beer Production}} \\ \hline
% & LP-SVQR      & 3    & 0.01   & 32  & 5   \\ 
% & ESVQR        & 2    & 20.00  & --  & 13  \\ 
% & LS-SVR       & 1    & 0.02   & --  & 128 \\ \hline
%\end{tabular}%
%}
%\end{table}

\subsection{ Numerical Results of  SVM based Conformal Regression method}
We now evaluate the performance of the SVM-based CR model (SVQR+CR) against the NN-based CQR model (CQR-NN) (\cite{romano2019conformalized}) on several benchmark datasets under the split conformal setting, with the target coverage level \(1 - \alpha = 0.90\). Both models were trained across 10 independent runs using fixed hyperparameter settings and identical splits of training and calibration sets.

The numerical results in Table~\ref{confromal} yield several insights. First, SVQR+CR achieved the target coverage in 4 out of 5 datasets, while CQR-NN did so in only 3 out of 5. Moreover, SVQR+CR yielded lower MPIW in 4 out of 5 cases. 

Most notably, the standard deviations of  PICP and MPIW across the 10 runs were zero for SVQR+CR, indicating perfectly stable predictions. In contrast, CQR-NN exhibited significant fluctuations. This is because, due to the non-convex nature of their optimization landscape, NN models often converge to different local minima across different training trials, even when the training data and hyperparameter settings remain fixed. Another key advantage of SVQR+CR is its reduced training time compared to CQR-NN, making it a more efficient and stable choice for conformal regression task.
\begin{table}[h]
  \centering
  {\fontsize{9}{10} \selectfont
  \begin{threeparttable}
    \begin{tabular}{|l|lcccccc|}
      \toprule
      \multirow{2}{*}{Dataset}
        & \multirow{2}{*}{Method}
        & \multicolumn{3}{c}{Performance}
        & \multicolumn{3}{c}{Std.\ Dev.\ (CQR-NN)} \\
      \cmidrule(lr){3-5} \cmidrule(lr){6-8}
      & & PICP (\%) & MPIW & Time (s) & PICP & MPIW & Time (s) \\
      \midrule
      Boston   & CQR-NN   & 94.12 & 2.19  & 2.65 & (1.14) & (0.12) & (0.23) \\
               & SVQR+CP  & 95.10 & 2.24  & 0.44 & (0.00) & (0.00) & (0.03) \\ \cline{1-8}
      Energy   & CQR-NN   & 87.66 & 1.13  & 3.20 & (1.12) & (0.03) & (0.31) \\
               & SVQR+CP  & 88.96 & 1.05  & 0.95 & (0.00) & (0.00) & (0.05) \\ \cline{1-8}
      Concrete & CQR-NN   & 92.58 & 19.62 & 1.89 & (1.02) & (0.71) & (0.17) \\
               & SVQR+CP  & 91.35 & 18.74 & 0.38 & (0.00) & (0.00) & (0.01) \\ \cline{1-8}
      Yacht    & CQR-NN   & 91.69 & 2.43  & 1.44 & (1.49) & (0.35) & (0.12) \\
               & SVQR+CP  & 90.82 & 2.87  & 0.22 & (0.00) & (0.00) & (0.01) \\ \cline{1-8}
      Servo    & CQR-NN   & 88.39 & 0.73  & 1.15 & (2.21) & (0.10) & (0.09) \\
               & SVQR+CP  & 89.74 & 0.68  & 0.16 & (0.00) & (0.00) & (0.01) \\ 
      \bottomrule
    \end{tabular}
    \caption{ Comprehensive comparison of CQR-NN and SVQR+CP.}
      \label{confromal}
  \end{threeparttable}}
\end{table}

\newpage 

\section{Future Work}
This paper presents a comprehensive roadmap for exploring UQ methods within the SVM regression framework. In contrast to NN models, SVM solutions exhibit lower uncertainty due to their tendency to converge to globally optimal solutions.

We proposed a feature selection algorithm tailored for PI estimation under the assumption that the bounds are linear functions of the input features. However, extending this approach to handle non-linear dependencies remains an important direction for future work, particularly in both NN and SVM-based models. Additionally, we show that SVM-based probabilistic forecasting models offer a compelling alternative to complex deep learning architectures by significantly reducing model complexity through the optimization of fewer parameters on batch datasets. Despite these advantages, they are not well-suited for dynamic or online data scenarios. To address this limitation, developing incremental or online SVM-based probabilistic forecasting models presents a promising avenue for future research. In this work, we have focused exclusively on the SVM regression model, a similar UQ analysis can be extended to SVM classification models in future research.


\subsubsection*{Acknowledgments}
I shall be thankful to Shayam Saktawat, Nikhil Bhanose and Chaitanya Dahale for helping in experiments.

 \bibliography{cas-refs}

\begin{thebibliography}{40}
\providecommand{\natexlab}[1]{#1}
\providecommand{\url}[1]{\texttt{#1}}
\expandafter\ifx\csname urlstyle\endcsname\relax
  \providecommand{\doi}[1]{doi: #1}\else
  \providecommand{\doi}{doi: \begingroup \urlstyle{rm}\Url}\fi

\bibitem[Anand et~al.(2020)Anand, Rastogi, and Chandra]{anand2020new}
Pritam Anand, Reshma Rastogi, and Suresh Chandra.
\newblock A new asymmetric $\epsilon$-insensitive pinball loss function based support
  vector quantile regression model.
\newblock \emph{Applied Soft Computing}, 94:\penalty0 106473, 2020.

\bibitem[Anand et~al.(2024)Anand, Bandyopadhyay, and Chandra]{anand2024tube}
Pritam Anand, Tathagata Bandyopadhyay, and Suresh Chandra.
\newblock Tube loss: A novel approach for prediction interval estimation and
  probabilistic forecasting.
\newblock \emph{arXiv preprint arXiv:2412.06853}, 2024.

\bibitem[Bishop(1994)]{bishop1994mixture}
Christopher~M Bishop.
\newblock Mixture density networks.
\newblock \emph{Neural Computing \& Applications}, 1994.

\bibitem[Bishop(1995)]{bishop1995neural}
Christopher~M Bishop.
\newblock Neural networks for pattern recognition.
\newblock \emph{Clarendon Press google schola}, 2:\penalty0 223--228, 1995.

\bibitem[Cannon(2011)]{cannon2011quantile}
Alex~J Cannon.
\newblock Quantile regression neural networks: Implementation in r and
  application to precipitation downscaling.
\newblock \emph{Computers \& geosciences}, 37\penalty0 (9):\penalty0
  1277--1284, 2011.

\bibitem[Cheng et~al.(2014)Cheng, Tezcan, and Cheng]{cheng2014confidence}
Qiang Cheng, Jale Tezcan, and Jie Cheng.
\newblock Confidence and prediction intervals for semiparametric mixed-effect
  least squares support vector machine.
\newblock \emph{Pattern Recognition Letters}, 40:\penalty0 88--95, 2014.

\bibitem[De~Brabanter et~al.(2010)De~Brabanter, De~Brabanter, Suykens, and
  De~Moor]{de2010approximate}
Kris De~Brabanter, Jos De~Brabanter, Johan~AK Suykens, and Bart De~Moor.
\newblock Approximate confidence and prediction intervals for least squares
  support vector regression.
\newblock \emph{IEEE Transactions on Neural Networks}, 22\penalty0
  (1):\penalty0 110--120, 2010.

\bibitem[De~VlEAUX et~al.(1998)De~VlEAUX, Schumi, Schweinsberg, and
  Ungar]{de1998prediction}
Richard~D De~VlEAUX, Jennifer Schumi, Jason Schweinsberg, and Lyle~H Ungar.
\newblock Prediction intervals for neural networks via nonlinear regression.
\newblock \emph{Technometrics}, 40\penalty0 (4):\penalty0 273--282, 1998.

\bibitem[He et~al.(2019)He, Qin, Wang, Wang, and Wang]{he2019electricity}
Yaoyao He, Yang Qin, Shuo Wang, Xu~Wang, and Chao Wang.
\newblock Electricity consumption probability density forecasting method based
  on lasso-quantile regression neural network.
\newblock \emph{Applied energy}, 233:\penalty0 565--575, 2019.

\bibitem[Hwang \& Ding(1997)Hwang and Ding]{hwang1997prediction}
JT~Gene Hwang and A~Adam Ding.
\newblock Prediction intervals for artificial neural networks.
\newblock \emph{Journal of the American Statistical Association}, 92\penalty0
  (438):\penalty0 748--757, 1997.

\bibitem[Kennedy \& Eberhart(1995)Kennedy and Eberhart]{kennedy1995particle}
James Kennedy and Russell Eberhart.
\newblock Particle swarm optimization.
\newblock In \emph{Proceedings of ICNN'95-international conference on neural
  networks}, volume~4, pp.\  1942--1948. ieee, 1995.

\bibitem[Khosravi et~al.(2010)Khosravi, Nahavandi, Creighton, and
  Atiya]{khosravi2010lower}
Abbas Khosravi, Saeid Nahavandi, Doug Creighton, and Amir~F Atiya.
\newblock Lower upper bound estimation method for construction of neural
  network-based prediction intervals.
\newblock \emph{IEEE transactions on neural networks}, 22\penalty0
  (3):\penalty0 337--346, 2010.

\bibitem[Koenker \& Bassett~Jr(1978)Koenker and
  Bassett~Jr]{koenker1978regression}
Roger Koenker and Gilbert Bassett~Jr.
\newblock Regression quantiles.
\newblock \emph{Econometrica: journal of the Econometric Society}, pp.\
  33--50, 1978.

\bibitem[Lauret et~al.(2017)Lauret, David, and Pedro]{lauret2017probabilistic}
Philippe Lauret, Mathieu David, and Hugo~TC Pedro.
\newblock Probabilistic solar forecasting using quantile regression models.
\newblock \emph{energies}, 10\penalty0 (10):\penalty0 1591, 2017.

\bibitem[MacKay(1992)]{mackay1992evidence}
David~JC MacKay.
\newblock The evidence framework applied to classification networks.
\newblock \emph{Neural computation}, 4\penalty0 (5):\penalty0 720--736, 1992.

\bibitem[Mercer(1909)]{mercer1909xvi}
James Mercer.
\newblock Xvi. functions of positive and negative type, and their connection
  the theory of integral equations.
\newblock \emph{Philosophical transactions of the royal society of London.
  Series A, containing papers of a mathematical or physical character},
  209\penalty0 (441-458):\penalty0 415--446, 1909.

\bibitem[Nix \& Weigend(1994)Nix and Weigend]{nix1994estimating}
David~A Nix and Andreas~S Weigend.
\newblock Estimating the mean and variance of the target probability
  distribution.
\newblock In \emph{Proceedings of 1994 ieee international conference on neural
  networks (ICNN'94)}, volume~1, pp.\  55--60. IEEE, 1994.

\bibitem[Papadopoulos(2008)]{papadopoulos2008inductive}
Harris Papadopoulos.
\newblock Inductive conformal prediction: Theory and application to neural
  networks.
\newblock In \emph{Tools in artificial intelligence}. Citeseer, 2008.

\bibitem[Papadopoulos et~al.(2002)Papadopoulos, Proedrou, Vovk, and
  Gammerman]{papadopoulos2002inductive}
Harris Papadopoulos, Kostas Proedrou, Volodya Vovk, and Alex Gammerman.
\newblock Inductive confidence machines for regression.
\newblock In \emph{Machine learning: ECML 2002: 13th European conference on
  machine learning Helsinki, Finland, August 19--23, 2002 proceedings 13}, pp.\
   345--356. Springer, 2002.

\bibitem[Pasche \& Engelke(2024)Pasche and Engelke]{pasche2024neural}
Olivier~C Pasche and Sebastian Engelke.
\newblock Neural networks for extreme quantile regression with an application
  to forecasting of flood risk.
\newblock \emph{The Annals of Applied Statistics}, 18\penalty0 (4):\penalty0
  2818--2839, 2024.

\bibitem[Pearce et~al.(2018)Pearce, Brintrup, Zaki, and Neely]{pearce2018high}
Tim Pearce, Alexandra Brintrup, Mohamed Zaki, and Andy Neely.
\newblock High-quality prediction intervals for deep learning: A
  distribution-free, ensembled approach.
\newblock In \emph{International conference on machine learning}, pp.\
  4075--4084. PMLR, 2018.

\bibitem[Rastogi et~al.(2018)Rastogi, Pal, and Chandra]{rastogi2018generalized}
Reshma Rastogi, Aman Pal, and Suresh Chandra.
\newblock Generalized pinball loss svms.
\newblock \emph{Neurocomputing}, 322:\penalty0 151--165, 2018.

\bibitem[Romano et~al.(2019)Romano, Patterson, and
  Candes]{romano2019conformalized}
Yaniv Romano, Evan Patterson, and Emmanuel Candes.
\newblock Conformalized quantile regression.
\newblock \emph{Advances in neural information processing systems}, 32, 2019.

\bibitem[Salinas et~al.(2020)Salinas, Flunkert, Gasthaus, and
  Januschowski]{salinas2020deepar}
David Salinas, Valentin Flunkert, Jan Gasthaus, and Tim Januschowski.
\newblock Deepar: Probabilistic forecasting with autoregressive recurrent
  networks.
\newblock \emph{International journal of forecasting}, 36\penalty0
  (3):\penalty0 1181--1191, 2020.

\bibitem[Sch{\"o}lkopf et~al.(2001)Sch{\"o}lkopf, Herbrich, and
  Smola]{scholkopf2001generalized}
Bernhard Sch{\"o}lkopf, Ralf Herbrich, and Alex~J Smola.
\newblock A generalized representer theorem.
\newblock In \emph{International conference on computational learning theory},
  pp.\  416--426. Springer, 2001.

\bibitem[Seber \& Seber(2015)Seber and Seber]{seber2015nonlinear}
George Seber and George~AF Seber.
\newblock Nonlinear regression models.
\newblock \emph{The Linear Model and Hypothesis: A General Unifying Theory},
  pp.\  117--128, 2015.

\bibitem[Shrivastava et~al.(2014)Shrivastava, Khosravi, and
  Panigrahi]{shrivastava2014prediction}
Nitin~Anand Shrivastava, Abbas Khosravi, and Bijaya~Ketan Panigrahi.
\newblock Prediction interval estimation for electricity price and demand using
  support vector machines.
\newblock In \emph{2014 International Joint Conference on Neural Networks
  (IJCNN)}, pp.\  3995--4002. IEEE, 2014.

\bibitem[Shrivastava et~al.(2015)Shrivastava, Khosravi, and
  Panigrahi]{shrivastava2015prediction}
Nitin~Anand Shrivastava, Abbas Khosravi, and Bijaya~Ketan Panigrahi.
\newblock Prediction interval estimation of electricity prices using pso-tuned
  support vector machines.
\newblock \emph{IEEE Transactions on Industrial Informatics}, 11\penalty0
  (2):\penalty0 322--331, 2015.

\bibitem[Suykens et~al.(2002)Suykens, De~Brabanter, Lukas, and
  Vandewalle]{suykens2002weighted}
Johan~AK Suykens, Jos De~Brabanter, Lukas Lukas, and Joos Vandewalle.
\newblock Weighted least squares support vector machines: robustness and sparse
  approximation.
\newblock \emph{Neurocomputing}, 48\penalty0 (1-4):\penalty0 85--105, 2002.

\bibitem[Takeuchi et~al.(2006)Takeuchi, Le, Sears, Smola, and
  Williams]{takeuchi2006nonparametric}
Ichiro Takeuchi, Quoc~V Le, Timothy~D Sears, Alexander~J Smola, and Chris
  Williams.
\newblock Nonparametric quantile estimation.
\newblock \emph{Journal of machine learning research}, 7\penalty0 (7), 2006.

\bibitem[Tanveer et~al.(2021)Tanveer, Sharma, Rastogi, and
  Anand]{tanveer2021sparse}
Mohammad Tanveer, Sweta Sharma, Reshma Rastogi, and Pritam Anand.
\newblock Sparse support vector machine with pinball loss.
\newblock \emph{Transactions on Emerging Telecommunications Technologies},
  32\penalty0 (2):\penalty0 e3820, 2021.

\bibitem[Taylor(2000)]{taylor2000quantile}
James~W Taylor.
\newblock A quantile regression neural network approach to estimating the
  conditional density of multiperiod returns.
\newblock \emph{Journal of forecasting}, 19\penalty0 (4):\penalty0 299--311,
  2000.

\bibitem[Vapnik(2013)]{vapnik2013nature}
Vladimir Vapnik.
\newblock \emph{The nature of statistical learning theory}.
\newblock Springer science \& business media, 2013.

\bibitem[Vovk et~al.(2005)Vovk, Gammerman, and Shafer]{vovk2005algorithmic}
Vladimir Vovk, Alexander Gammerman, and Glenn Shafer.
\newblock \emph{Algorithmic learning in a random world}, volume~29.
\newblock Springer, 2005.

\bibitem[Vovk et~al.(1999)Vovk, Gammerman, and Saunders]{vovk1999machine}
Volodya Vovk, Alexander Gammerman, and Craig Saunders.
\newblock Machine-learning applications of algorithmic randomness.
\newblock 1999.

\bibitem[Wan et~al.(2016)Wan, Lin, Wang, Song, and Dong]{wan2016direct}
Can Wan, Jin Lin, Jianhui Wang, Yonghua Song, and Zhao~Yang Dong.
\newblock Direct quantile regression for nonparametric probabilistic
  forecasting of wind power generation.
\newblock \emph{IEEE Transactions on Power Systems}, 32\penalty0 (4):\penalty0
  2767--2778, 2016.

\bibitem[Ye et~al.(2025)Ye, Wang, and Chen]{ye2025nonlinear}
Ya-Fen Ye, Jie Wang, and Wei-Jie Chen.
\newblock Nonlinear feature selection for support vector quantile regression.
\newblock \emph{Neural Networks}, 185:\penalty0 107136, 2025.

\bibitem[Zhang et~al.(2020{\natexlab{a}})Zhang, Liu, Yan, Han, Li, and
  Long]{zhang2020improved}
Hao Zhang, Yongqian Liu, Jie Yan, Shuang Han, Li~Li, and Quan Long.
\newblock Improved deep mixture density network for regional wind power
  probabilistic forecasting.
\newblock \emph{IEEE Transactions on Power Systems}, 35\penalty0 (4):\penalty0
  2549--2560, 2020{\natexlab{a}}.

\bibitem[Zhang et~al.(2018)Zhang, Quan, and Srinivasan]{zhang2018improved}
Wenjie Zhang, Hao Quan, and Dipti Srinivasan.
\newblock An improved quantile regression neural network for probabilistic load
  forecasting.
\newblock \emph{IEEE Transactions on Smart Grid}, 10\penalty0 (4):\penalty0
  4425--4434, 2018.

\bibitem[Zhang et~al.(2020{\natexlab{b}})Zhang, Quan, Gandhi, Rajagopal, Tan,
  and Srinivasan]{zhang2020improving}
Wenjie Zhang, Hao Quan, Oktoviano Gandhi, Ram Rajagopal, Chin-Woo Tan, and
  Dipti Srinivasan.
\newblock Improving probabilistic load forecasting using quantile regression nn
  with skip connections.
\newblock \emph{IEEE Transactions on Smart Grid}, 11\penalty0 (6):\penalty0
  5442--5450, 2020{\natexlab{b}}.

\end{thebibliography}
 \bibliographystyle{tmlr}

%\appendix
%\section{Appendix}
%You may include other additional sections here.

\end{document}